%% file: sep_filled.tex
\pdfoutput=1 
\documentclass[11pt]{article}

\usepackage[font=libertinus, citestyle=authoryear]{kurbanlab}

\input{affiliations}   

\usepackage{threeparttable}
\usepackage{pgfplots}
\pgfplotsset{compat=1.18}
\usepgfplotslibrary{fillbetween}
\usetikzlibrary{positioning,arrows.meta,calc,patterns,decorations.pathreplacing}
\definecolor{cNarrow}{HTML}{2B6CB0}
\definecolor{cWide}{HTML}{C05621}
\definecolor{cBand}{HTML}{ED8936}
\definecolor{cInk}{HTML}{2D3748}
\definecolor{cGrid}{HTML}{CBD5E0}
\definecolor{cMute}{HTML}{718096}
\definecolor{cFill}{HTML}{EDF2F7}
\definecolor{cRes}{HTML}{9B2C2C}

\tikzset{
  fig/.style     = {font=\small, color=cInk, line cap=round, line join=round},
  panel/.style   = {font=\scriptsize\bfseries, color=cInk},
  lbl/.style     = {font=\scriptsize, color=cInk},
  note/.style    = {font=\scriptsize, color=cMute},
  rule/.style    = {draw=cInk, line width=0.5pt},
  grid/.style    = {draw=cGrid, line width=0.4pt},
  cut/.style     = {draw=cInk, dash pattern=on 2pt off 1.6pt, line width=0.6pt},
  bar/.style     = {draw=cMute, line width=0.4pt, fill=cInk!10},
  barboth/.style = {draw=cNarrow, line width=0.5pt, fill=cNarrow!18},
  bardrop/.style = {draw=cGrid, line width=0.4pt, fill=cInk!6},
  barhi/.style   = {draw=cWide, line width=0.4pt, fill=cBand, fill opacity=0.55,
                    draw opacity=0.9},
  tok/.style     = {rounded corners=2pt, minimum height=5.2mm,
                    minimum width=8.4mm, inner sep=1pt, font=\scriptsize,
                    line width=0.5pt},
  tokS/.style    = {tok, draw=cMute, fill=cFill},
  tokA/.style    = {tok, draw=cNarrow, fill=cNarrow!10},
  tokB/.style    = {tok, draw=cWide,   fill=cWide!12},
  flow/.style    = {draw=cMute, line width=0.7pt,
                    -{Latex[length=1.6mm,width=1.3mm]}},
  lead/.style    = {draw=cMute, line width=0.45pt,
                    -{Latex[length=1.2mm,width=1.0mm]}},
}

\pgfplotsset{
  seplot/.style = {
    width=6.2cm, height=4.0cm,
    axis line style={draw=cInk, line width=0.5pt},
    tick label style={font=\scriptsize, color=cInk},
    label style={font=\scriptsize, color=cInk},
    title style={font=\scriptsize\bfseries, color=cInk, yshift=-1pt},
    legend style={font=\scriptsize, draw=cGrid, fill=white,
                  inner sep=1.5pt, row sep=-1pt},
    grid=major, major grid style={draw=cGrid, line width=0.4pt},
    tick align=outside, tick style={color=cMute}, axis on top,
  }
}

\newcommand{\yes}{\ensuremath{\checkmark}}
\newcommand{\swept}{\ensuremath{\bullet}}
\newcommand{\no}{\ensuremath{\times}}
\newcommand{\rowgap}{\addlinespace[2pt]}

\newcommand{\Sep}{\mathrm{Sep}}
\newcommand{\TV}{\mathrm{TV}}
\newcommand{\Ex}{\mathbb{E}}
\newcommand{\Prob}{\mathbb{P}}
\newcommand{\Vocab}{\mathcal{V}}
\newcommand{\ind}[1]{\mathbf{1}\!\left[#1\right]}

\title{Truncation Samplers Are Nested: Comparing Decoding Rules Is a Measurement Problem}
\RunningTitle{Truncation samplers are nested}

\Author{Rasul Khanbayov}{hbku}
\Author[corresponding=hkurban@hbku.edu.qa, orcid=0000-0003-3142-2866]{Hasan Kurban}{hbku}

\CodeURL{https://github.com/KurbanIntelligenceLab/truncation-samplers}

\begin{document}
\maketitle

\begin{abstract}
Settling one comparison between two truncation samplers has cost six
thousand A100-hours, and nothing says in advance how far apart two samplers
can be. We observe that almost every deployed truncation rule keeps a prefix
of the probability-sorted vocabulary, so any two such rules are nested at
every decoding step. Nesting yields a closed-form per-step distance and a
coupling under which both samplers emit identical text until they first
disagree. Accumulated over a generation, this gives a separation certificate
that bounds the difference between two samplers on any bounded outcome, is
computable from logs a comparison already produces, and equals their
sequence-level total variation distance exactly when neither rule ever
overtakes the other. Two consequences follow without measurement: exactness
is fragile, and certifying a small difference requires the two retained
masses to agree to very high precision at every step. On six open-weight
models over four math and science benchmarks, the certificate for every
hyperparameter grid we tested is at least $0.998$, so a bound alone rarely
settles a comparison, although two members of one family can be far closer.
What remains useful is the coupling, which gives an unbiased paired
estimator that cuts variance $1.2$--$3.8\times$ against prompt-level
pairing. Read against benchmark sizes at the default pair's variance and
our point estimate of its between-prompt component, resolving a one-point
difference needs $8$ to $32$ samples per prompt on GSM8K, MATH-500 and
GPQA-Diamond and about $210$ on AIME 2025; at that component's upper
confidence bound GPQA-Diamond needs more than $32$ and AIME 2025 cannot be
resolved at any sample count. Choosing a truncation
sampler is a measurement problem at least as much as a design one. The code
is available at \url{https://github.com/KurbanIntelligenceLab/truncation-samplers}.
\end{abstract}

\section{Introduction}
\label{sec:intro}

Training-free samplers decide how to draw from a language model's next-token
distribution, and the dominant family truncates. Top-$k$ keeps a fixed number
of tokens \citep{fan2018hierarchical}, top-$p$ a fixed cumulative mass
\citep{holtzman2020curious}, and a further series of rules from the last
five years adapt the cutoff to the shape of the distribution
\citep{hewitt2022truncation,basu2021mirostat,nguyen2025minp,tang2025topnsigma,%
pless2026,toph2026,topb2026,ants2026,mink2026,topw2026}. Roughly a dozen are
in use, and choosing among them is expensive and contested. A
re-examination of min-$p$ found its reported advantage does not survive once
swept hyperparameters are equalized across samplers \citep{schaeffer2025minp},
at a cost of six thousand A100-hours, and that cost is not specific to
one paper. Reasoning benchmarks swing by double digits across seeds
\citep{soberlook2025}, benchmark variance is under-reported
\citep{madaan2024variance,biderman2024lessons}, and run-to-run variation can
misrank systems \citep{hiddenerror2026}.

Two things make these comparisons harder than they need to be. First,
nothing says in advance how far apart two truncation rules can be, so a
study cannot be sized before it is run and a reported gap cannot be
checked against what the samplers could have produced. Second, every
protocol that makes two rules comparable does so by equalizing a scalar
summary, whether hyperparameter grid volume \citep{schaeffer2025minp},
estimated risk level \citep{zhou2024balancing}, or expected retained mass
\citep{kempton2025lnd}. To our knowledge none has been shown to be the
quantity that governs how far apart the rules can be, so a matched
comparison rests on an untested assumption.

Most samplers above, all but Top-W and locally typical sampling, retain a
\emph{prefix} of the vocabulary sorted by probability, so at any decoding
step the candidate sets of any two are nested, meaning one contains the
other. The narrower rule's distribution is then exactly the wider rule's
conditioned on the narrower set, so the distance between them is a ratio
of two numbers decoding already computes. The two rules can also be run so
they emit identical tokens until the first draw falls in the band between
the cutoffs (Figure \ref{fig:teaser}). That coupling accumulates the
per-step distance into a bound on any bounded outcome, computable from
logs a comparison already produces. Written in log form it also answers
the second problem, because it is a dispersion while the matching
criterion is a mean.

\begin{center}
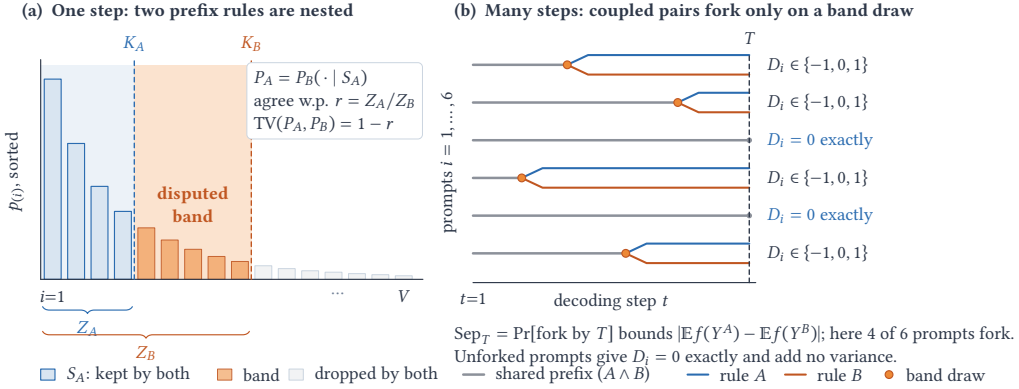

\begin{minipage}{\linewidth}
\centering
\scalebox{0.86}{\input{figures/sep_fig_teaser}}
\captionof{figure}{\textbf{Nesting gives a free per-step distance and a
coupling.} \textbf{(a)} Two prefix rules cut the sorted vocabulary at
$K_A\le K_B$ and disagree only on the shaded band. Drawing from $B$, both
emit the draw if it lands in $S_A$, otherwise $A$ redraws, so they agree
with probability $r=Z_A/Z_B$. \textbf{(b)} Coupled pairs share tokens until
a draw lands in the band, then fork. An unforked prompt gives
$D_i=f(Y^A_i)-f(Y^B_i)=0$ exactly, and the fraction that forks estimates
$\Sep_T$.}
\label{fig:teaser}
\end{minipage}
\end{center}

Measured on real models, the certificate is large. It is at least $0.998$
between the narrowest and widest members of every hyperparameter grid we
tested, which is expected once a grid contains untruncated and near-greedy
members, and it stays large between rules matched on operating point,
which matching does not prevent, since it equalizes a mean while separation
is governed by a dispersion. Under the maximal coupling such samplers
diverge on almost every generation, and for the grid extremes, which never
flip, this is also their exact total variation distance. Their accuracies
barely differ: on the eight pairs where we measured both, every accuracy
gap is at most $2.4\%$ of its certificate. A bound
cannot settle such a comparison in advance, so it has to be measured, and
nesting helps there too. The same coupling gives a paired estimator whose
variance is bounded by the certificate, and reading its measured variances
against benchmark sizes shows which benchmarks can resolve a one-point
difference at all. Choosing a truncation sampler is a measurement problem at
least as much as a design one.

\paragraph{Contributions.} (1) We show that prefix truncation rules are
nested, which gives an exact per-step distance and a rejection-free maximal
coupling between any two of them (Lemma \ref{lem:nest}). (2) Accumulating
it gives the \emph{separation certificate}, which bounds the difference on
any bounded outcome, is exact iff no trajectory flips which rule is
narrower, and covers an entire hyperparameter grid from one decoding run
(Theorem \ref{thm:cert}, Corollary \ref{cor:grid}). (3) Two structural
results say when a certificate can be exact or small: exactness decays
geometrically once the narrower rule can change (Proposition
\ref{prop:fragile}), and a sufficient per-step budget with a dispersion
lower bound bracket when a small certificate is possible (Corollary
\ref{cor:budget}). (4) Measured on seven open-weight models (six for the
grid, plus a distilled reasoner for long horizons), the certificate is large on every
grid we tested, the coupling cuts estimator variance $1.2$--$3.8\times$, and
the resulting sample-size arithmetic shows which benchmarks can resolve a
one-point difference (Sections \ref{sec:experiments} and
\ref{sec:benchmarks}).

\section{Prefix Truncation Rules}
\label{sec:setup}

Fix a vocabulary $\Vocab$ of size $V$. At step $t$ with context $x_{<t}$ the
model gives $p_t \in \Delta^{V-1}$; write $p_{(1)} \ge \cdots \ge p_{(V)}$ for
the probabilities in descending order.

\begin{definition}[Prefix rule]
\label{def:prefix}
A \emph{prefix rule} maps a context and a next-token distribution to a cutoff
index $K(x_{<t},p_t)\in\{1,\dots,V\}$, retains $S_K=\{(1),\dots,(K)\}$ and
samples from $P_K(i)=p_{(i)}/Z_K$, where $Z_K=\sum_{i\le K}p_{(i)}$ is the
\emph{retained mass}. The order is sorted by descending probability and
breaks ties by token index, and all rules under comparison must use the
same order.
\end{definition}

Two prefixes of one total order are always nested, so for any prefix rules
$A,B$ and any step, $S_A\subseteq S_B$ or $S_B\subseteq S_A$, and every
result below rests on this. Table \ref{tab:rules} writes the deployed
samplers in this form.

Comparing rules also needs a common notion of aggressiveness, since any
sampler can behave like greedy or untruncated decoding and raw
hyperparameters are not comparable across families. Following
\citet{kempton2025lnd}, we match on the operating point $\rho(K)=\Ex[Z_K]$,
the expected retained mass over a teacher-forced calibration corpus of 200
held-out GSM8K prompts (Appendix \ref{app:expdetails}). Rules are
operating-matched at level $\rho$ when their hyperparameters give equal
operating points.

\begin{table}[t]
\centering
\begin{threeparttable}
\caption{\textbf{Most deployed truncation samplers return a cutoff index in
the sorted order,} so the retained set is a prefix and any two such rules
are nested at every step; Top-W and locally typical sampling, marked below
the rule, do not. The nine rows above the gap are the family used in
Section \ref{sec:experiments}: six marked \swept\ each contribute six
hyperparameter values, and three marked single (basic, $p$-less and Top-H,
none swept over multiple settings) contribute one member each.}
\label{tab:rules}
\small
\setlength{\tabcolsep}{5pt}
\renewcommand{\arraystretch}{0.98}
\begin{tabular}{@{}llcc@{}}
\toprule
\textbf{Sampler} & \textbf{Cutoff rule} & \textbf{Prefix} & \textbf{Swept} \\
\midrule
basic (no truncation)                    & $K=V$                                              & \yes & single \\
top-$k$ \citep{fan2018hierarchical}      & $K=k$                                              & \yes & \swept \\
top-$p$ \citep{holtzman2020curious}      & smallest $K$ with $Z_K\ge p$                       & \yes & \swept \\
$\epsilon$ \citep{hewitt2022truncation}  & keep $p_{(i)} \ge \epsilon$                        & \yes & \swept \\
$\eta$ \citep{hewitt2022truncation}      & keep $p_{(i)} \ge \min\{\epsilon,\sqrt{\epsilon}e^{-H}\}$ & \yes & \swept \\[1pt]
min-$p$ \citep{nguyen2025minp}           & keep $p_{(i)} \ge p_{\mathrm{base}}\,p_{(1)}$      & \yes & \swept \\
top-$n\sigma$ \citep{tang2025topnsigma}  & keep $\ell_{(i)} \ge \ell_{(1)}-n\,\mathrm{sd}(\ell)$ & \yes & \swept \\
$p$-less \citep{pless2026}               & collision entropy, no hyperparameter               & \yes & single \\
Top-H \citep{toph2026}                   & bounded-entropy cutoff at its default              & \yes & single \\
\rowgap
min-$k$ \citep{mink2026}                 & index at the local logit-decay cliff               & \yes & \\
mirostat \citep{basu2021mirostat}        & index from a perplexity controller                 & \yes & \\
\midrule
top-W \citep{topw2026}                   & prefix of a scan over embedding geometry, not $p$  & \no & \\
locally typical \citep{meister2023typical} & orders by $|{-}\log p_{(i)}-H|$ & \no & \\
\bottomrule
\end{tabular}
\begin{tablenotes}[flushleft]\footnotesize
\item Every rule above the gap returns a prefix of the sorted order, most by
comparing a threshold against $p_{(i)}$, which is monotone in $i$; min-$k$
and mirostat return an index directly. Across
$997{,}465{,}872$ logged cutoffs there are zero out-of-range cutoffs and
zero non-nested pairs (mirostat, not implemented, is classified from its
description). Top-W orders tokens by a score that mixes embedding geometry
with log-probability, and locally typical sampling omits the top token on
$8.8\%$ of our logged steps at $\tau=0.2$ (Appendix \ref{app:audit}).
\end{tablenotes}
\end{threeparttable}
\end{table}

\section{Separation Theory}
\label{sec:theory}

Nesting becomes a distance, the distance becomes a bound, and the coupling
behind the bound becomes an estimator. Section \ref{sec:exact} derives the
certificate for a pair and for a whole grid, Section \ref{sec:small} asks
when it can be exact or small, and Section \ref{sec:estimator} turns the
coupling into an estimator. Proofs are in Appendix \ref{app:proofs}.

\subsection{Nesting gives an exact distance}
\label{sec:exact}

Write $A,B$ for two prefix rules acting on the same $p_t$, with retained
masses $Z_A,Z_B$ and $r=\min\{Z_A,Z_B\}/\max\{Z_A,Z_B\}$, defined by a
minimum and a maximum since which rule is narrower can change from step
to step.

\begin{lemma}[Nesting]
\label{lem:nest}
At any step, $S_A \subseteq S_B$ or $S_B \subseteq S_A$. Relabel so
$S_A \subseteq S_B$. Then
\emph{(i)} $P_A = P_B(\,\cdot \mid S_A)$, hence
$\TV(P_A,P_B) = 1 - Z_A/Z_B = 1-r$; and
\emph{(ii)} drawing $X\sim P_B$ and setting $Y=X$ if $X\in S_A$, else drawing
$Y\sim P_A$ afresh, gives marginals $(P_A,P_B)$ with $\Prob(Y\neq X)=1-r$, so
this coupling is maximal.
\end{lemma}

\begin{figure}[t]
\centering
\includegraphics[width=\linewidth]{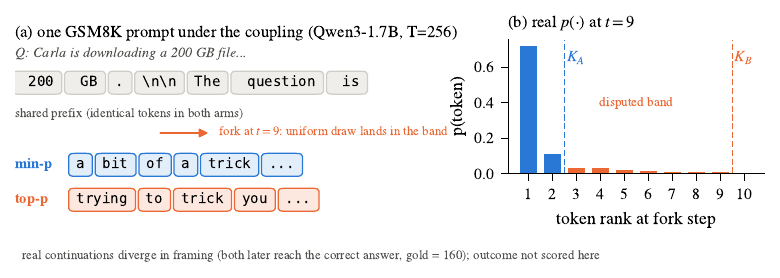}
\caption{\textbf{The coupling on one GSM8K prompt}
(Qwen3-1.7B, min-$p$ versus top-$p$, $T{=}256$). \textbf{(a)} The rules share
tokens until step $9$, where the draw lands in the band: top-$p$ keeps it,
min-$p$ redraws, and both continuations later reach the correct answer.
\textbf{(b)} The next-token distribution at the fork, with each rule's
cutoff.}
\label{fig:coupledexample}
\end{figure}

The whole distance lives in the disputed band between the cutoffs. Part (ii)
is the standard maximal coupling for nested supports
\citep{lindvall2002lectures}: draw from the wider rule, and if the draw
lands in the narrower set both samplers emit it, otherwise the narrower
rule redraws and the two continue independently (Figure
\ref{fig:coupledexample} shows one such fork on a real prompt). Because the
narrower rule can be $A$ at some steps and $B$ at others, the shared prefix
follows the \emph{pointwise-narrower} rule $A\wedge B$,
$K_{A\wedge B}=\min\{K_A,K_B\}$, not $A$.

\begin{theorem}[Separation certificate]
\label{thm:cert}
Let $f$ map a generated sequence to $[0,1]$ and set
\[
\Sep_T(A,B) \;=\; 1 - \Ex_{y \sim A \wedge B}\Big[\textstyle\prod_{t=1}^{T} r_t(y_{<t})\Big].
\]
\emph{(i)} $\Sep_T$ is the probability that a coupled pair diverges within $T$
steps, and $|\Ex f(Y^A) - \Ex f(Y^B)| \le \Sep_T$; for $f$ bounded in $[a,b]$
the bound is $(b-a)\Sep_T$.
\emph{(ii)} $\TV(P^{(T)}_A,P^{(T)}_B) \le \Sep_T$, with equality iff
no trajectory in the common support ever flips which rule is narrower.
\end{theorem}

Part (i) is the instrument, one number that no bounded metric can exceed;
part (ii) says when it is tight rather than merely valid, which
Proposition \ref{prop:fragile} makes precise. Computing it needs only the
retained masses of both rules along one $A\wedge B$ walk per prompt
(Appendix \ref{app:expdetails}), so it is available from any decoding
run's own logs. The same construction extends from a pair to a whole
grid.

\begin{corollary}[One number for a whole grid]
\label{cor:grid}
Let $\mathcal{G}$ be a finite family of prefix rules, for instance every
sampler in a study crossed with its hyperparameter grid, and let
$\mathcal{G}^{\wedge},\mathcal{G}^{\vee}$ be its pointwise-narrowest and
pointwise-widest bounding rules, both prefix rules but not necessarily
members of $\mathcal{G}$ itself. All of $\mathcal{G}$ can be
coupled simultaneously, so for every $A,B \in \mathcal{G}$ and every $f$
bounded in $[0,1]$,
$|\Ex f(Y^A) - \Ex f(Y^B)| \le \Sep_T(\mathcal{G}^{\wedge},\mathcal{G}^{\vee})
=: \Sep_T(\mathcal{G})$, computed by the same procedure above with
$A=\mathcal{G}^{\wedge}$, $B=\mathcal{G}^{\vee}$. When $\mathcal{G}$ is a
chain under pointwise inclusion its extremes are themselves members, so no
smaller scalar bounds every pair (Appendix \ref{app:extras} gives the metric
that attains it).
\end{corollary}

Coupling the family at once is what yields a common scalar, since bounding
each pair separately would put every pair under its own measure. One run
under $\mathcal{G}^{\wedge}$ gives $\Sep_T(\mathcal{G})$, but certifying an
individual pair still needs a run under that pair's own $A\wedge B$ or
under one of its members (Proposition \ref{prop:onerule}), since
$\mathcal{G}^{\wedge}$'s trajectories do not cover the region the pairwise
expectation integrates over.

\subsection{When can a certificate be exact or small?}
\label{sec:small}

The log form connects the certificate to existing theory. Write
$g_t = \log Z_{A,t} - \log Z_{B,t}$ for the per-step log-gap between the two
rules' retained masses. With
$\Delta_T = -\Ex_{A \wedge B}[\sum_t \log r_t]$,
\begin{equation}
\Delta_T = \Ex_{A \wedge B}\Big[\textstyle\sum_{t\le T} |g_t|\Big]
\;\ge\; \Ex_{A \wedge B}\Big[\big|\textstyle\sum_{t\le T} g_t\big|\Big],
\label{eq:lnd}
\end{equation}
with equality iff no trajectory flips. The right side is the difference of
the two rules' \emph{log} local normalization distortions
\citep{kempton2025lnd} (for a single rule, $\prod_t Z_t^{-1}$), so
$\Delta_T$ is the accumulated \emph{absolute} log-gap while theirs is the
\emph{net} log-gap, and the two coincide exactly when the certificate is
exact. Equation \eqref{eq:lnd} is also a dispersion, whereas the operating
point of Section \ref{sec:setup} is a mean, and equalizing a mean does not
constrain a spread.

\begin{proposition}[Exactness is fragile]
\label{prop:fragile}
Write $s_t=\mathrm{sign}(g_t)\in\{-1,0,+1\}$ along an $A\wedge B$ trajectory
and call step $t$ \emph{decisive} if $s_t\neq0$. A trajectory \emph{flips}
if it has decisive steps of both signs; ties never break constancy, so
Theorem \ref{thm:cert}(ii) is exact precisely when no trajectory in the
common support flips. Suppose that from every reachable prefix, $s_{t+1}=+1$
and $s_{t+1}=-1$ each have conditional probability at least $\varepsilon>0$
(ties may absorb the rest of the mass). Then
$\Prob(\text{no flip through } T) \le 2(1-\varepsilon)^T-(1-2\varepsilon)^T$,
and this is sharp.
\end{proposition}

Exactness therefore survives only when one rule stays weakly narrower
throughout on every trajectory in the common support, as in a monotone
sweep within one family. Any pair whose order can change loses it
geometrically in the horizon ($3.1\times10^{-23}$ at $\varepsilon=0.05$,
$T=1024$). Exactness and fairness pull against each other: matching
operating points equalizes $\Ex[Z]$ across the two rules, exactly the
condition that makes neither sign dominant.

\begin{corollary}[Per-step budget]
\label{cor:budget}
Let $S=\sum_{t\le T}|g_t|$ along an $A\wedge B$ trajectory, so $\Delta_T=\Ex[S]$.
\emph{(a)} If $\Delta_T\le-\log(1-\gamma)$ then $\Sep_T\le\gamma$: a
time-averaged per-step gap of $\frac1T\sum_{t\le T}\Ex[|g_t|]\le
-\log(1-\gamma)/T$ suffices to certify $\Sep_T\le\gamma$, with no
assumption that the per-step terms are identically distributed.
\emph{(b)} For every $s>0$, $\Sep_T\ge(1-e^{-s})\Prob(S\ge s)$; taking
$s=\Delta_T/2$ and applying Chebyshev's inequality gives
$\Sep_T\ge\big(1-e^{-\Delta_T/2}\big)\big(1-4\Var(S)/\Delta_T^2\big)$.
\end{corollary}

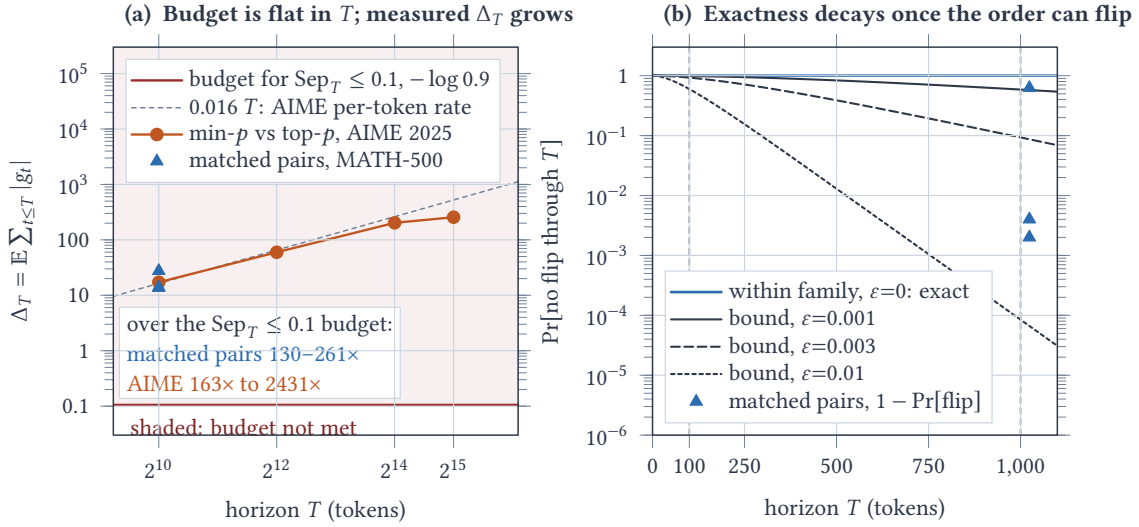
\begin{figure}[t]
\centering
\resizebox{\linewidth}{!}{\input{figures/sep_fig_theory}}
\caption{\textbf{Measurements against the two structural
results.} Markers are measured, solid lines are theory. \textbf{(a)} The
sufficient budget of Corollary \ref{cor:budget}(a) for $\Sep_T\le0.1$ is flat
in $T$; measured $\Delta_T$ exceeds it $130$--$261\times$ for the matched
pairs of Table \ref{tab:matched} (triangles, $T{=}1024$) and
$163$--$2431\times$ for the AIME series (circles; gray dashes extend its
per-token rate, $0.016\,T$). \textbf{(b)} Proposition \ref{prop:fragile}'s
bound on the no-flip probability for three $\varepsilon$, against the
matched pairs' measured $1-\Pr[\text{flip}]$ at $T{=}1024$. Within a family
($\varepsilon{=}0$) exactness holds at every horizon; at
$\varepsilon{=}0.05$ the bound is $3.1\times10^{-23}$ by $T{=}1024$.}
\label{fig:theory}
\end{figure}

Part (a) is sufficient but does not explain why a certificate is large,
since $\Delta_T$ can vastly exceed the budget while $\Sep_T$ stays small if
$S$ is heavy-tailed. Part (b) closes that gap: when $S$ concentrates around
its mean, $\Sep_T$ itself must be large. The budget of part (a) is
demanding. Ruling out a $0.1$ difference needs the two retained masses to
agree to a mean relative gap of $1.0\times10^{-4}$ per step over $1{,}024$
tokens, and $3.2\times10^{-6}$ over $32{,}768$. Retained masses sit near
$1$ and move with context, so two different rules rarely hold that
precision for a thousand consecutive steps, and nothing we measured comes
within two orders of magnitude of the budget (Figure \ref{fig:theory}).

\subsection{From coupling to estimator}
\label{sec:estimator}

\begin{proposition}[Coupled estimator]
\label{prop:var}
Let $f$ map a generated sequence to $[0,1]$ and $D_i = f(Y^A_i)-f(Y^B_i)$
over $n$ coupled pairs. Then
$\hat\Delta=\frac1n\sum_i D_i$ is unbiased for $\Ex f(Y^A)-\Ex f(Y^B)$, the
same estimand independent runs target, and $\Var(D) \le \Ex[D^2] \le \Sep_T$.
Write $\sigma_A^2=\Var(f(Y^A))$ and $\sigma_B^2=\Var(f(Y^B))$ for the
outcome variances under independent sampling from each rule. When $\Sep_T>0$,
$\Var(D_{\mathrm{indep}})/\Var(D_{\mathrm{coupled}}) \ge
(\sigma_A^2+\sigma_B^2)/\Sep_T$.
\end{proposition}

Every prompt where the pair never diverges contributes an exact zero, so
the estimator pays variance only where the samplers disagree. This is the
common-random-numbers argument \citep{glasserman1992crn} applied to a
coupling nesting makes explicit; what is new is that $\Sep_T$, computable
from the comparison's own logs or, by Proposition \ref{prop:onerule}, from
a single rule's logs, bounds the variance ratio without a separate
calibration run. The guarantee holds against independent runs, and
prompt-level pairing is a stronger baseline for which we claim no theory.
The ratio bound exceeds $1$ only when $\sigma_A^2+\sigma_B^2>\Sep_T$, so
for a binary outcome ($\sigma_A^2+\sigma_B^2\le1/2$) it guarantees no
improvement once $\Sep_T>1/2$.

\paragraph{Assumptions.} Both rules must be prefix rules acting on the same
$p_t$ under the same sorted order, so temperature and every other logit
processor is shared; every experiment here uses temperature $1.0$, applied
before truncation (unequal temperatures: Appendix \ref{app:extras}).
Outcomes are bounded and white-box decoding control is required. A maximum
over a swept comparison is a multiple comparison, bounded by Corollary
\ref{cor:grid} but still needing correction when reporting intervals.

\section{Experiments}
\label{sec:experiments}

We check the harness against Theorem \ref{thm:cert}, then ask how large the
certificate is, whether it is exact, whether the coupling reduces variance,
and whether it tracks the real gap.

\paragraph{Setup.} Open-weight models a single GPU can serve: Qwen3 at 0.6B,
1.7B, 4B and 8B \citep{qwen3}, Gemma 3 4B \citep{gemma3} and gpt-oss-20b
\citep{gptoss} for an architecture contrast, and
DeepSeek-R1-Distill-Qwen-1.5B \citep{deepseekr1} for long chain of thought.
Outcomes are automatically verifiable, so no judge enters the measurement:
MATH-500 \citep{hendrycks2021math}, GPQA-Diamond \citep{rein2023gpqa},
GSM8K \citep{cobbe2021gsm8k}, and AIME 2025 with the generation cap at
$32{,}768$ tokens. Six of the nine samplers marked in Table \ref{tab:rules}
are each swept over six log-spaced values, an equal-size grid following
the hyperparameter-volume point of \citet{schaeffer2025minp}, and the other
three contribute one member each, giving a grid $\mathcal{G}$ of 39 members
and $741$ pairs. All decoding, including every estimator arm, runs in one
custom loop over HuggingFace Transformers \citep{wolf2020transformers} at
batch size one, since next-token logits drift with batch size.
Certificates use the top $2048$ logged probabilities and exclude steps
whose cutoff would exceed that, an exclusion the grid certificate never
needs. All decoding totals $64.5$ H200-GPU-hours, under two percent of the
six thousand A100-hours of \citet{schaeffer2025minp}. Appendix
\ref{app:expdetails} gives the full grid, the exclusion analysis and the
compute accounting.

Apart from the gate and the five-seed estimator results, every quantity
below comes from a single configuration without a reported interval, so
those comparisons are descriptive rather than inferential, and we name
each confound where it appears.

\subsection{Results}
\label{sec:results}

\paragraph{Does the harness measure what the theorem says?} Theorem
\ref{thm:cert}(i) predicts the certificate equals the coupled divergence
rate, so the harness can be checked before it is trusted. On
Qwen3-1.7B/GSM8K (min-$p{=}0.05$ versus top-$p{=}0.95$, $n{=}200$) the
certificate is $0.722$ against a measured divergence rate of $0.73$
(95\% CI $[0.67,0.79]$) at $T{=}256$, and $0.945$ against $0.935$ (CI
$[0.90,0.965]$) at $T{=}1024$. The certificate falls inside the bootstrap
interval at both horizons, and three controls pass (Appendix
\ref{app:results}).

\paragraph{How large is the certificate?} Across nine model and task
configurations (Qwen3 0.6B/4B/8B on GSM8K, Qwen3-1.7B on GSM8K,
GPQA-Diamond, MATH-500 and AIME 2025, and Gemma~3~4B and gpt-oss-20b on
MATH-500) the grid certificate $\Sep_T(\mathcal{G})$ is at least $0.998$
($0.998064$ at the minimum, $95\%$ CI $[0.997832,0.998296]$ over seeds),
falling on Qwen3-1.7B/GSM8K, and above $0.9997$ on every other
configuration (Figure \ref{fig:gridcert}, Appendix \ref{app:results}). Two qualifications matter. The
maximum comes from comparisons against $p$-less and against wide-top-$p$
and small-$\epsilon$ endpoints, so this is a statement about
hyperparameter-volume-controlled sweeps rather than about truncation rules
in general. And on the seven Qwen3 configurations $K_{\mathcal{G}^{\wedge}}$
equals one on $99.3\%$ of $9{,}006{,}080$ logged steps, so the certificate
is measured along essentially greedy paths rather than typical
trajectories of any single member rule. Rules matched on operating
point stay far apart too (Table \ref{tab:matched}), consistent with
Corollary \ref{cor:budget}(a): their retained masses miss the sufficient
budget by two orders of magnitude. Within a family the picture differs, as
the budget allows: a monotone top-$k$ sweep from $10$ to $320$ gives
certificates as low as $0.10$.

\paragraph{Is the certificate exact?} Exactness decides whether the number
is a property of the samplers or an artifact of the bound, and Theorem
\ref{thm:cert}(ii) makes it checkable from the same logs. Within a family
the flip rate is exactly $0$ everywhere, so exactness holds in practice,
while across families it is $0.15$--$0.36$ on raw grids, and $0.998$,
$0.996$ and $0.374$ for the three matched pairs of Table \ref{tab:matched}.
Proposition \ref{prop:fragile} predicts this ordering, since matching
makes neither order dominant. We read the contrast as consistent with that
prediction rather than establishing it, because the raw and matched rates
come from different pairs and possibly different horizons, and a flip rate
is interpretable only at a stated horizon (Figure \ref{fig:theory}b). A
sweep over five matched operating points shows a moderate relationship in
the direction Equation \eqref{eq:lnd} predicts (Appendix \ref{app:results}).

\paragraph{Longer generations make it worse.} Every step of a long
reasoning chain is another chance to diverge. On DeepSeek-R1-Distill-Qwen-1.5B/AIME~2025 with the default pair
(min-$p{=}0.05$, top-$p{=}0.95$), read at four horizons as prefixes of one
$T{=}32{,}768$ run, $\Delta_T$ is $17.2$ at $T{=}1024$, $59.7$ at $4096$,
$202.8$ at $16384$ and $256.2$ at $32768$, and $\Sep_T$, computed directly
over the $30$ prompts, is $0.9988$ at $T{=}1024$ and rounds to $1.00$ at
the three longer horizons. It is already above $0.5$ at the shortest cap,
the falsification condition specified in advance for the certificate's
usefulness as a smallness guarantee. At $T{=}32{,}768$ the per-step budget
for a $0.1$ certificate is $3.2\times10^{-6}$, and $\Delta_T$ exceeds it by
a factor of $2431$. Since most generations finish before the longer caps,
Appendix \ref{app:results} also reports $\Delta_T$ per realized token.

\paragraph{Does the coupling reduce variance?} The coupling is useful
whether the certificate is large or small. On GSM8K, a single
run at each of four model scales gave variance reductions of $1.2\times$
(Qwen3-0.6B), $1.8\times$ (4B), $1.9\times$ (8B) and $2.1\times$ (1.7B)
relative to prompt-level pairing, and Table \ref{tab:variance} reports the
five-seed and cross-benchmark results. The GPQA-Diamond estimate is large
but not significant, likely because near-guessing accuracy destabilizes a
variance ratio. None of the five-seed pairs reaches $\Sep_T<1/2$ (the
lowest is $0.71$), so Proposition \ref{prop:var}'s guarantee does not apply
and these reductions are empirical; two simpler ways of sharing randomness
fall significantly short of the maximal coupling (Appendix
\ref{app:results}). A negative control, top-$p$ $0.98$ versus top-$p$
$0.99$ ($\Sep_T=0.437$), is the pair where the guarantee applies: with
$\sigma_A^2=0.240$ and $\sigma_B^2=0.241$ it guarantees only $1.10\times$,
while the measured reduction is about $12.0\times$, so the guarantee is far
from tight there.

\begin{table}[t]
\centering
\begin{threeparttable}
\caption{\textbf{The coupled estimator's variance reduction against
prompt-level pairing.} GSM8K rows are Qwen3-1.7B with five seeds, cluster
bootstrap intervals and Holm correction across the three pairs; the
MATH-500 and GPQA-Diamond rows replicate the default pair.}
\label{tab:variance}
\small
\setlength{\tabcolsep}{6pt}
\begin{tabular}{@{}llcl@{}}
\toprule
\textbf{Task} & \textbf{Pair} & \textbf{Reduction} & \textbf{95\% CI, significance} \\
\midrule
GSM8K        & min-$p$(0.05) vs top-$p$(0.95)      & $3.8\times$  & $[3.1,4.9]$, significant after Holm \\
GSM8K        & top-$k$(10) vs top-$p$(0.70)        & $1.3\times$  & $[1.1,1.5]$, significant after Holm \\
GSM8K        & $\epsilon$(0.1) vs $\eta$(0.0003)   & $1.3\times$  & $[1.2,1.6]$, significant after Holm \\
MATH-500     & min-$p$(0.05) vs top-$p$(0.95)      & $1.7\times$  & $[1.2,2.4]$, bootstrap $p=0.023$ \\
GPQA-Diamond & min-$p$(0.05) vs top-$p$(0.95)      & $18.7\times$ & $[8.0,62.1]$, bootstrap $p=0.16$ \\
\bottomrule
\end{tabular}
\end{threeparttable}
\end{table}

\begin{figure}[t]
\centering
\includegraphics[width=\linewidth]{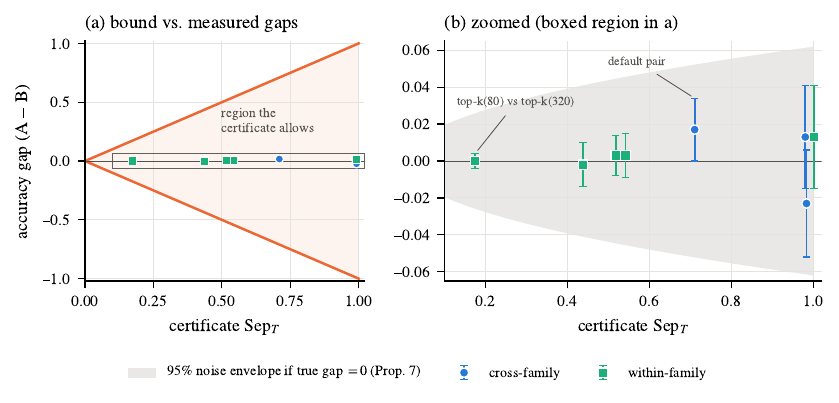}
\caption{\textbf{Certificates are large, accuracy gaps
are small} (the eight pairs of Table \ref{tab:predictive}). \textbf{(a)}
Theorem \ref{thm:cert} bounds the expected gap by the certificate (cone).
\textbf{(b)} The boxed region with $95\%$ CIs. The gray band,
$\pm1.96\sqrt{\Sep_T/n}$ with $n{=}1{,}000$, is the zero-gap spread implied
by Proposition \ref{prop:var}; it is conservative in using a variance bound
and anti-conservative in treating $200$ prompts $\times$ five seeds as
independent. Points near $\Sep_T{=}0.99$ are offset by $\pm0.009$.}
\label{fig:certgap}
\end{figure}

\paragraph{Does the certificate predict the gap?} A bound can be correct
and still useless if it never tracks the real gap. Table
\ref{tab:predictive} (Appendix \ref{app:results}) lists eight pairs on
Qwen3-1.7B/GSM8K at $T{=}256$, three cross-family and five within-family
including the negative control, with $\Sep_T$ from $0.17$ to $0.99$. Every
measured gap sits far inside its certificate (Figure \ref{fig:certgap}; a
tightness ratio, gap over $\Sep_T$, of at most $0.024$). On
GSM8K no generation stops before the cap, so part of the measured
divergence is text written after the answer, which the grader sees only
through a conflicting second answer (Appendix \ref{app:extras}). $\Sep_T$
correlates with the absolute gap at Pearson $r=0.85$ ($95\%$ CI
$[0.35,0.97]$), but this is not on its own evidence of prediction:
Proposition \ref{prop:var} gives $\Var(D)\le\Sep_T$, so pure sampling noise
with every true gap equal to zero already induces a positive correlation.
Against a simulated null that draws each pair's gap from a zero-mean
Gaussian at its own measured standard error, the null mean of $r$ is $0.54$
and the observed $r=0.85$ sits at its $94.5$th percentile (one-sided
$p=0.055$); standardizing each gap by its own standard error still gives
$r=0.68$ against a null mean of $0$ (two-sided $p=0.056$). Both sit at the
conventional significance boundary, so at eight pairs we read the result
as inconclusive, neither a demonstration nor a null artifact. Whether a
small certificate ever appears between two \emph{cross-family} rules
remains open.

\section{What This Means for Benchmark Comparisons}
\label{sec:benchmarks}

The estimator's variances have a consequence that follows from the coupled
arm alone. At one sample per prompt, detecting a one-point difference between
two samplers needs roughly $787$--$16{,}379$ prompts under coupling and
$14{,}686$--$23{,}727$ under prompt-level pairing, across the pairs of
Table \ref{tab:variance}. Both lower ends come from GPQA-Diamond, whose
$18.7\times$ reduction is not significant, so the representative reduction
is the $1.3$--$3.8\times$ of the other four pairs. At one sample per prompt
every benchmark, AIME 2025 included, falls short of at least one measured
requirement.

Drawing $k$ samples per prompt changes this. For the default pair on GSM8K
the between-prompt variance $\tau^2$ is statistically indistinguishable
from zero (bootstrap $95\%$ upper bound $9.3\times10^{-4}$ over prompts), so
the required prompt count falls close to proportionally with $k$ (Figure
\ref{fig:samplesize}, Appendix \ref{app:results}). At the point estimate, GSM8K reaches the requirement
by $k{=}8$, MATH-500 by $k{=}32$, and GPQA-Diamond almost exactly at
$k{=}32$, though not at $\tau^2$'s upper bound. AIME 2025 is still
$6.6\times$ short at $k{=}32$ and would need $k\approx210$, while at the
upper bound no $k$ suffices, and this extrapolation assumes the $\tau^2$
measured on GSM8K rather than on AIME 2025. At the default pair's
variance, coupling plus resampling removes the constraint for the larger
benchmarks; whether it
does for the smallest one is not settled by what we measured (Appendix
\ref{app:results}).

\paragraph{What to report.} Three quantities cost one decoding run each:
every sampler's operating point with its flip rate, $\Sep_T$ for each
compared pair beside the measured gap, flagging any gap above it, and the
per-step retained masses, which make every certificate recomputable.

\section{Related Work}
\label{sec:related}

\paragraph{Local normalization distortion.} The closest prior work is
\citet{kempton2025lnd}, who isolate a \emph{local normalization
distortion}, the reciprocal of the product of retained masses along a
generated sequence, show it pulls decoding away from the quality-diversity
frontier, and find that its variance across sequences rather than its mean
is what matters. Their matching criterion is the operating point we adopt,
and their comparison is between a rule and its own globally normalized
counterpart. Ours is between two deployed rules, and Equation
\eqref{eq:lnd} shows the two coincide exactly when no trajectory flips.

\paragraph{Truncation rules and matched comparisons.}
\citet{holtzman2020curious} established truncation as the remedy for
degeneration and \citet{hewitt2022truncation} gave it a generative reading;
most later rules differ only in which statistic sets the cutoff. Among
protocols that match samplers on a scalar summary,
\citet{zhou2024balancing} is closest in spirit, matching at equal estimated
risk levels, but without coupling, so obtaining no bound on an end-task
difference. Existing theory explains why truncation helps rather than how
much two truncations differ
\citep{hewitt2022truncation,finlayson2024closing,decodinggame2024}.

\paragraph{Coupling.} Coupling through shared randomness to reduce variance
is standard
\citep{glasserman1992crn,yang1991crn,lindvall2002lectures,bavarian2020correlated};
what is new is not the construction but that truncation rules make the
maximal coupling closed-form and rejection-free at every step. Pairing on
shared prompts is established practice
\citep{mcnemar1947note,dror2018hitchhiker} and documented to be
underpowered \citep{card2020power}. Estimating sequence-level total
variation between autoregressive distributions is generally hard
\citep{price2026tvestimation}; ours is a structured case where it is
computed rather than estimated. Appendix \ref{app:relwork} covers
speculative decoding, which on a nested pair reduces to exactly this
coupling, and further related lines.

\section{Limitations and Conclusion}

\paragraph{Limitations.} The theory covers prefix rules on a shared base
distribution, excluding locally typical sampling, Top-W, contrastive
decoding, reordering rules and cross-model comparison; a certificate rules
differences out, never in, and both certificate and estimator degrade as
the divergence probability approaches one. Sequence-level total variation
is estimable from a run under either rule alone (Proposition
\ref{prop:onerule}) but has not yet been computed on a real log, so
Theorem \ref{thm:cert}(ii)'s equality is checked directly only on synthetic
models, with measured flip rates as indirect evidence. The grid's $741$
pairwise certificates are not yet decoded. The predictive-validity check
uses eight GSM8K pairs on one model and horizon, five within-family, and
its null-calibrated test lands at the significance boundary. Every task is
short-answer mathematics or science, whereas truncation samplers were
designed for open-ended generation, so our certificates may be large
partly because reasoning chains are not the regime these rules target,
and on GSM8K, where no generation stops before the cap, part of each
certificate reflects text written after the answer.

\paragraph{Conclusion.} Nesting makes the distance between two truncation
samplers free to compute. The certificate is large while accuracy gaps are
small, so a comparison has to be measured rather than bounded. The useful
question is not only which truncation rule is best but whether a given
experiment, on a given benchmark, was ever sized to tell.

\begin{kilrepro}
Every result in Section \ref{sec:theory} is proved in Appendix
\ref{app:proofs}, with its assumptions stated alongside it, and the
supporting results in Appendix \ref{app:extras} are stated with their
sources and are checked numerically where applicable. The code repository,
\url{https://github.com/KurbanIntelligenceLab/truncation-samplers},
contains the reference implementation of every quantity defined in
Section \ref{sec:theory} and the two scripts behind the checks of Appendix
\ref{app:verification}, both of which run on CPU with NumPy alone. Section
\ref{sec:experiments} names models, datasets and statistical procedures.
All datasets are public and all models are open-weight. Code and per-step
retained-mass logs are available in the same repository, and both are
runnable from a clean checkout with the commands in
\texttt{REPRODUCE.md}. Model checkpoints are pinned to a specific revision,
not \texttt{main}, for every model reported, Qwen3-0.6B/1.7B/4B/8B,
Gemma~3~4B, gpt-oss-20b and DeepSeek-R1-Distill-Qwen-1.5B. GSM8K (MIT), GPQA-Diamond (CC-BY-4.0),
AIME-2025 (Apache-2.0) and MATH-500 (MIT, via the underlying Hendrycks
et al. MATH dataset) are each pinned to a specific dataset revision.
Every generation uses five seeds ($0$--$4$) where
seeds are reported at all, and one fixed prompt template per task with no
chat formatting, few-shot examples or system prompt, so the certificate is
measured against the model's base per-step distribution. Where a
falsification condition is stated for a result, such as the threshold in
Section \ref{sec:results} for the certificate's usefulness as a smallness
guarantee, it was specified in advance of running the corresponding
experiment, not registered with an external, timestamped service, and we
say ``specified in advance'' rather than ``pre-registered'' for that
reason.
\end{kilrepro}




\section*{Ethics Statement}
This work involves no human subjects, no personal data and no model
training. It studies evaluation methodology for existing open-weight
models on public benchmarks. One consideration deserves note. An audit
tool applied to published work can be misused to overstate criticism, and
a certificate bounds what two samplers can differ by but does not
establish that any method lacks value.

\section*{Use of Generative AI}
Generative AI tools were used substantially throughout this work: in
framing the theory and formulating and proving its claims, in implementing
the code and running the experiments, in interpreting results, in
reviewing related literature, and in drafting and editing the text. Every
formal claim is checked by the synthetic verification suite of Appendix
\ref{app:verification}, which caught several errors during preparation,
and a human author independently re-derived Lemma \ref{lem:nest}, Theorem
\ref{thm:cert} and Proposition \ref{prop:var} and checked the coupled
decoder against Lemma \ref{lem:nest}(ii). Every number about a real
language model comes from a logged decoding run. The human authors take
responsibility for the final content of this work.

\bibliography{references}

\appendix
\section{Proofs}
\label{app:proofs}

Throughout, at the step in question $K_A \le K_B$ without loss of generality,
so $S_A \subseteq S_B$, $Z_A=\sum_{i\le K_A}p_{(i)}$,
$Z_B=\sum_{i\le K_B}p_{(i)}$ and $r=Z_A/Z_B$. Both rules act on the same $p$
under the same sorted order. Since $K\ge1$ and $p_{(1)}>0$, every retained mass
is strictly positive.

\paragraph{Lemma \ref{lem:nest}, nesting.}
$S_A$ and $S_B$ are prefixes of the same total order, so the one with the
smaller cutoff index is contained in the other. $\square$

\paragraph{Lemma \ref{lem:nest}(i).}
For $i \in S_A$, $P_B(i \mid S_A)=P_B(i)/P_B(S_A)=(p_{(i)}/Z_B)/(Z_A/Z_B)
=p_{(i)}/Z_A=P_A(i)$, and both sides vanish off $S_A$. For the distance,
\begin{align*}
2\,\TV(P_A,P_B)
&= \sum_{i \in S_A}\Big|\tfrac{p_{(i)}}{Z_A}-\tfrac{p_{(i)}}{Z_B}\Big|
 + \sum_{i \in S_B\setminus S_A}\tfrac{p_{(i)}}{Z_B}
= Z_A\Big(\tfrac{1}{Z_A}-\tfrac{1}{Z_B}\Big) + \tfrac{Z_B-Z_A}{Z_B}
= 2\Big(1-\tfrac{Z_A}{Z_B}\Big),
\end{align*}
using $Z_A \le Z_B$ to drop the absolute values. $\square$

\paragraph{Lemma \ref{lem:nest}(ii).}
The $B$-marginal holds by construction. For $i \in S_A$,
$\Prob(Y=i)=\Prob(X=i)+\Prob(X\notin S_A)P_A(i)
=\tfrac{p_{(i)}}{Z_B}+(1-\tfrac{Z_A}{Z_B})\tfrac{p_{(i)}}{Z_A}
=\tfrac{p_{(i)}}{Z_A}=P_A(i)$, and $\Prob(Y=i)=0$ off $S_A$. When
$X\in S_A$ we set $Y=X$, and when $X\notin S_A$ we have $Y\in S_A$ and
$X\notin S_A$, so $Y\ne X$ necessarily. Hence
$\Prob(Y\ne X)=\Prob(X\notin S_A)=1-r=\TV(P_A,P_B)$ exactly, and since $\TV$
is the minimum of $\Prob(Y\ne X)$ over all couplings, this one is maximal.
$\square$

\paragraph{Theorem \ref{thm:cert}(i).}
Run the coupling of Lemma \ref{lem:nest}(ii) at every step. Given agreement
through $t-1$, Lemma
\ref{lem:nest}(ii) gives agreement at step $t$ with probability $r_t(y_{<t})$,
and by Lemma \ref{lem:nest}(i) the emitted token is then distributed under the
narrower rule at that step, which is $P_{A\wedge B}$. Since the narrower
rule may be $A$ at some steps and $B$ at others, the shared prefix is
distributed as an $A\wedge B$ trajectory and not in general as an $A$
trajectory. Chaining,
$\Prob(\text{no divergence in }T)=\Ex_{A\wedge B}[\prod_{t\le T} r_t]$, so the
divergence probability equals $\Sep_T$. Each arm follows its own rule
marginally at every step, by Lemma \ref{lem:nest}(ii) before divergence and by
construction after, so $Y^A$ and $Y^B$ have the laws of full $A$- and
$B$-trajectories. On the no-divergence event $Y^A=Y^B$, hence
$|\Ex f(Y^A)-\Ex f(Y^B)|=|\Ex[(f(Y^A)-f(Y^B))\ind{\text{div}}]| \le
(b-a)\Sep_T$ for $f\in[a,b]$. $\square$

\paragraph{Theorem \ref{thm:cert}(ii).}
First, $y$ lies in $\mathrm{supp}(A)\cap\mathrm{supp}(B)$ iff every $y_t$ lies
in $S_{A,t}\cap S_{B,t}=S_{A\wedge B,t}$, so the common support is exactly
$\mathrm{supp}(A\wedge B)$. The coupled sequences differ exactly on the
divergence event, so $\Prob(Y^A \ne Y^B)=\Sep_T$, and as
$\TV$ is the minimum of $\Prob(Y^A\ne Y^B)$ over couplings, $\TV \le \Sep_T$.

For equality, write $a_t=P_A(y_t\mid y_{<t})$ and $b_t=P_B(y_t\mid y_{<t})$, so
$A(y)=\prod_t a_t$, $B(y)=\prod_t b_t$, and
$\min\{a_t,b_t\}=p_{y_t}/Z_{\max,t}$ on the common support. Then
$1-\Sep_T = \sum_{y \in \mathrm{supp}(A\wedge B)} \prod_t \min\{a_t,b_t\}$ and
$1-\TV = \sum_y \min\{A(y),B(y)\}$, the latter having zero terms off the
common support. Termwise $\min\{\prod a_t,\prod b_t\} \ge \prod \min\{a_t,b_t\}$,
which re-proves $\TV \le \Sep_T$. Since every term of positive weight must
agree, fix $y$ and let $S=\{t:a_t<b_t\}$, $L=\{t:a_t>b_t\}$. If $L=\emptyset$
then $\prod\min=\prod a\le\prod b$ and the two agree, and symmetrically if
$S=\emptyset$. If both are nonempty then $\prod\min$ is strictly smaller
than both products, because it replaces $a_s$ by $b_s<a_s$ for $s\in L$ and
$b_t$ by $a_t<b_t$ for $t\in S$, all factors being strictly positive. Since
$a_t<b_t$ iff $Z_A>Z_B$, both sets nonempty is exactly a flip. $\square$

\paragraph{Corollary \ref{cor:grid}.}
Pointwise minima and maxima of finitely many prefix rules are prefix rules.
At a step, $S_{\mathcal{G}^{\vee}}$ equals the retained set of whichever
member attains the maximum, so drawing $X \sim P_{\mathcal{G}^{\vee}}$ is
realizable, every $R \in \mathcal{G}$ with $X \in S_R$ emits $X$, and every
other draws from its own $P_R$. Each marginal is correct by the computation
in Lemma \ref{lem:nest}(ii) with $\mathcal{G}^{\vee}$ in place of $B$, for
each $R$ separately and regardless of how the remaining draws are
correlated. All rules agree iff $X \in S_{\mathcal{G}^{\wedge}}$. If so all
emit $X$, and if not, the member attaining the pointwise minimum redraws
inside $S_{\mathcal{G}^{\wedge}}$ while some member emits
$X \notin S_{\mathcal{G}^{\wedge}}$. Conditional on agreement the common
token is $P_{\mathcal{G}^{\vee}}(\cdot\mid S_{\mathcal{G}^{\wedge}})
=P_{\mathcal{G}^{\wedge}}$ by Lemma \ref{lem:nest}(i), and the argument of
Theorem \ref{thm:cert}(i) applies. $\square$

\paragraph{Proposition \ref{prop:fragile}.}
Track two states along a trajectory that has not yet flipped: $U$, no
decisive step yet, and $D$, every decisive step so far shares one sign.
From $D$ with sign $c$, the next step flips the trajectory (ends it) with
conditional probability at least $\varepsilon$ (the opposite sign), so $D$
survives $n$ further steps with probability at most $(1-\varepsilon)^n$.
From $U$, the next step is decisive with some probability $q\ge2\varepsilon$
and, being the first decisive step, cannot itself flip anything; let $v_n$
be the largest probability of not having flipped after $n$ steps starting
from $U$. Conditioning on the first step,
$v_n=\max_{q\ge2\varepsilon}\big[q\,(1-\varepsilon)^{n-1}+(1-q)\,v_{n-1}\big]$,
since a decisive first step moves to $D$ with $n-1$ steps left to flip
(probability of surviving those $\le(1-\varepsilon)^{n-1}$, attained by
splitting $q$ evenly between the two signs so the survival bound is tight
regardless of which sign was drawn) and a tied first step stays in $U$ with
$n-1$ steps left. By induction $v_{n-1}\ge(1-\varepsilon)^{n-1}\ge0$, so the
bracket is non-increasing in $q$ (its $q$-derivative is
$(1-\varepsilon)^{n-1}-v_{n-1}\le0$) and the maximum over $q\ge2\varepsilon$
sits at the smallest feasible value, $q=2\varepsilon$, giving
$v_n=2\varepsilon(1-\varepsilon)^{n-1}+(1-2\varepsilon)v_{n-1}$ with
$v_0=1$. Unrolling this recursion gives the closed form
$v_T=2(1-\varepsilon)^T-(1-2\varepsilon)^T$, which one checks satisfies the
same recursion and boundary condition by direct substitution. The process
that puts mass exactly $\varepsilon$ on each sign while in $U$, and exactly
$\varepsilon$ on the opposite sign once in $D$, attains every inequality
above, so the bound is sharp. $\square$

\paragraph{Corollary \ref{cor:budget}.}
Here $\Sep_T=1-\Ex[e^{-S}]$.
\emph{(a)} $x\mapsto e^{-x}$ is convex, so Jensen gives
$\Ex[e^{-S}]\ge e^{-\Delta_T}$ and hence $\Sep_T\le1-e^{-\Delta_T}$.
Requiring the right side to be at most $\gamma$ gives
$\Delta_T\le-\log(1-\gamma)$, and since $\Delta_T=\Ex[S]=\sum_{t\le T}
\Ex[|g_t|]=T\cdot\frac1T\sum_{t\le T}\Ex[|g_t|]$ with no assumption that the
per-step terms are identically distributed, this is exactly a bound on
their time average, giving the per-step form. This bound is close to
tight for small $\gamma$, and at $\gamma=0.1$ the two sides differ by $5\%$.
\emph{(b)} Pointwise, $1-e^{-S}\ge(1-e^{-s})\ind{S\ge s}$ for every $s\ge0$,
since $1-e^{-S}\ge0$ and is increasing in $S$; taking expectations gives
$\Sep_T\ge(1-e^{-s})\Prob(S\ge s)$. At $s=\Delta_T/2$, Chebyshev's
inequality gives $\Prob(S<\Delta_T/2)\le\Prob(|S-\Delta_T|\ge\Delta_T/2)\le
4\Var(S)/\Delta_T^2$, so $\Prob(S\ge\Delta_T/2)\ge1-4\Var(S)/\Delta_T^2$,
which substituted above gives the stated bound. $\square$

\paragraph{Proposition \ref{prop:var}.}
Unbiasedness holds because Lemma \ref{lem:nest}(ii) preserves both marginals.
Since $D=0$ off the divergence event and $f\in[0,1]$, $|D|\le\ind{\text{div}}$,
so $\Var(D)\le\Ex[D^2]\le\Sep_T$. For independent runs
$\Var(f(Y^A)-f(Y^B))=\sigma_A^2+\sigma_B^2$, giving the ratio whenever
$\Sep_T>0$. $\square$

\paragraph{Proposition \ref{prop:onerule}.}
On $\mathrm{supp}(A)\cap\mathrm{supp}(B)$, the $A\wedge B$ walk agrees with
the $A$ walk exactly when every step lands in $S_{A\wedge B}=S_A\cap S_B$,
which for $y\in\mathrm{supp}(B)$ happens for every step of the $A$ walk
since $S_A\subseteq S_B$ or $S_B\subseteq S_A$ at each step (Lemma
\ref{lem:nest}). So, writing $r_t=\min(Z_{A,t},Z_{B,t})/\max(Z_{A,t},
Z_{B,t})$ as in Theorem \ref{thm:cert},
\[
(A\wedge B)(y)\prod_t r_t
=\prod_t\frac{p(y_t)}{\max(Z_{A,t},Z_{B,t})}
=A(y)\prod_t\frac{Z_{A,t}}{\max(Z_{A,t},Z_{B,t})}
\]
for $y\in\mathrm{supp}(A)\cap\mathrm{supp}(B)$, and both sides are $0$ for
$y\notin\mathrm{supp}(B)$ since then some step has $y_t\notin S_B$ so the
$A\wedge B$ walk could not have reached $y$. Summing over $y$ and using
Theorem \ref{thm:cert}(i)'s $1-\Sep_T=\sum_y(A\wedge B)(y)\prod_t r_t$
gives the first identity. For the second, $B(y)/A(y)=\prod_t
p(y_t)/Z_{B,t}\cdot Z_{A,t}/p(y_t)=\prod_t Z_{A,t}/Z_{B,t}$ on
$\mathrm{supp}(A)\cap\mathrm{supp}(B)$, wherever this ratio is defined, and
$B(y)=0$ for $y\in\mathrm{supp}(A)\setminus\mathrm{supp}(B)$; so
$\TV=\sum_y(A(y)-B(y))_+=\Ex_{y\sim A}[(1-B(y)/A(y))_+]$, which is the
stated form once $B(y)/A(y)$ is written out. Both integrands are products of
terms in $(0,1]$, truncated by an indicator or a positive part, so they lie
in $[0,1]$, and $A$ is the sampling measure in both expectations, so a
single run under $A$ estimates either quantity with the usual
Hoeffding guarantee for a bounded statistic. $\square$

\section{Supporting Results}
\label{app:extras}

Four facts Section \ref{sec:theory} refers to are recorded here briefly
rather than as full statements, since none is novel to this paper.
When $K_A(p)\le K_B(p)$ at every reachable context, the supremum in Theorem
\ref{thm:cert}(ii) is attained at $f^\star(y)=\ind{y \in \mathrm{supp}(A)}$,
which we verify by exact enumeration in Appendix \ref{app:verification}.
Unequal temperatures are covered by the standard construction for
non-nested but still totally ordered supports. If a shared logit
transformation is applied before truncation but the two samplers use
different temperatures, retained sets stay nested since temperature
preserves the sorted order, but Lemma \ref{lem:nest}(i)'s equality fails
because $P_A(i)/P_B(i)$ now depends on $i$, so the explicit no-rejection
coupling no longer applies directly. What survives is that the per-step
distance between the two distributions is still computable in closed form
at no extra query cost, and Theorem \ref{thm:cert} holds with $1-r_t$
replaced by that distance, with the expectation taken along the induced
overlap measure rather than along $A\wedge B$. When both rules
are driven by one shared uniform random variable through the inverse CDF
of their truncated distributions, as arithmetic and quasi-Monte-Carlo
samplers do \citep{vilnis2023arithmetic,li2026quasimotto}, the resulting
coupling is valid but not maximal, so its divergence probability is at
least $1-r$, though the certificate of Theorem \ref{thm:cert} still bounds
the outcome difference since that bound is a property of the marginals and
holds under any coupling. Finally, Theorem \ref{thm:cert} as stated fixes
$T$; with EOS, let $\tau$ be the step at which either rule first emits it
(inclusive), a stopping time with respect to the shared $A\wedge B$
trajectory since whether step $t$'s emitted token is EOS is observable at
step $t$. Replacing $T$ by $\tau\wedge T$ throughout, so the product stops
at $\tau\wedge T$ rather than running to a fixed $T$ regardless of EOS,
gives $\Sep_{\tau\wedge T}$ and the same two parts of Theorem
\ref{thm:cert} with the same proof, since the induction it relies on never
used $T$ being non-random, only that it upper-bounds the number of steps
taken. Our own certificate logs already use this stopping time, since
\texttt{decode.py} logs the EOS-emitting step and then stops (Section
\ref{sec:experiments} reports the resulting \texttt{eos\_rate} for the
long-horizon AIME series), so every $\Sep_T$ reported in this paper is
already $\Sep_{\tau\wedge T}$ in this sense. The variance-reduction
estimator's coupled, prompt-paired and independent arms do not currently
share this stopping rule: they run the full $T$ steps regardless of EOS,
and grading does not simply take the first or the last answer the
completion states. Math tasks are graded with \texttt{math\_verify}, which
parses every \texttt{\textbackslash boxed\{\}} expression in the
completion and scores it correct only if exactly one distinct value
appears and it matches the gold answer, so a completion that states an
answer, reconsiders, and states a different one anywhere later scores
incorrect regardless of which value is right; GPQA is graded by the first
standalone option letter found, which does take the first answer. So
outcome noise from text generated after a would-be stopping point is
real, but its effect is task-dependent rather than uniformly "extra
sampling after the answer is ignored." This is a real difference between the
certificate and the estimator, not merely a presentational gap, and is not
resolved here.

\begin{proposition}[Pairwise certificate and total variation from one rule's
run]
\label{prop:onerule}
For prefix rules $A$ and $B$ and $y\sim P_A^{(T)}$,
\begin{align*}
1-\Sep_T(A,B) &= \Ex_{y\sim A}\Big[\ind{y\in\mathrm{supp}(B)}
\prod_{t\le T}\frac{Z_{A,t}}{\max(Z_{A,t},Z_{B,t})}\Big],\\
\TV\big(P_A^{(T)},P_B^{(T)}\big) &= \Ex_{y\sim A}\Big[\Big(1-
\ind{y\in\mathrm{supp}(B)}\prod_{t\le T}\frac{Z_{A,t}}{Z_{B,t}}\Big)_+\Big].
\end{align*}
Both integrands lie in $[0,1]$, so a Monte Carlo estimate from a single
rule's decoding run admits a Hoeffding interval.
\end{proposition}

A run under $A$ alone therefore gives both the pairwise certificate and the
exact sequence-level total variation for every $B$ compared against it, not
only the pointwise-narrower member of a pair: a grid of $n$ swept rules
needs $n$ runs for all $\binom{n}{2}$ pairwise quantities in either
direction, rather than one run per pair under that pair's own
$A\wedge B$. This does not change $\Sep_T$ or $\TV$ themselves, only which
decoding run recovers them; the trajectory-level coupling of Theorem
\ref{thm:cert}(i) still needs the $A\wedge B$ walk, since it is the joint
law of $(Y^A,Y^B)$, not just their marginals' distance, that the
one-rule identity does not reconstruct. Applying it needs no new
\emph{sampling}, but it is not free of engineering cost: the estimator's
independent-arm runs in this paper (Section \ref{sec:experiments}) save
only the final token sequence, not the per-step retained masses
$Z_{A,t}$ the identity is stated in terms of, so recovering it for those
runs needs a further pass, either logging retained masses on a fresh
independent-arm decode or a teacher-forced forward pass over the
already-sampled sequences, not a re-read of data already on disk.

\section{Extended Related Work}
\label{app:relwork}

\paragraph{Evaluation of truncation.} \citet{kempton2025lnd} characterize
top-$k$, nucleus and temperature sampling as equilibrium states and show
the local normalization distortion is large. Nesting is what makes their
object usable for comparing two rules, because the intractable global
constant that forces \citet{kempton2025lnd} to work with paired ratios
cancels structurally and the maximal coupling becomes explicit. \citet{gareev2024localglobal}
compare local and global decoding empirically, again within a rule rather
than across rules. \citet{zhou2024balancing} match at equal estimated risk
levels via a prefix tree, comparing distributional quantities on
open-ended text. \citet{truncationblindspot2026} ask what truncation
removes relative to human text and \citet{ravfogel2023conformal} calibrate
top-$p$ per entropy bin, both distinct goals.

\paragraph{Speculative decoding.} On a nested pair, speculative decoding's
accept-or-resample-from-the-residual rule
\citep{leviathan2023speculative,listcoupling2025} reduces to exactly the
coupling of Lemma \ref{lem:nest}(ii): with $P_A=P_B(\cdot\mid S_A)$, the
acceptance probability $\min(1,P_A(x)/P_B(x))$ is $1$ on $S_A$ and $0$
outside it, so acceptance is the deterministic set membership Lemma
\ref{lem:nest}(ii) uses rather than a stochastic ratio, and the residual
$\propto\max(0,P_A-P_B)$ is $P_A$ itself. Speculative decoding's target
use, drafting from a cheap model to skip evaluations of an expensive one,
is unrelated to ours, comparing two truncation rules of the same model for
measurement, but the mechanism is the same textbook construction
specialized to nested supports rather than a new one; the contribution
here is showing that truncation rules make this coupling closed-form and
rejection-free at every step, not designing it.

\paragraph{Other couplings.} Adaptive parallel decoding accepts to first
disagreement \citep{israel2025adaptive}, and watermarking couples
generation to a key \citep{kuditipudi2024watermark}. A parallel line fixes
the truncation rule and redesigns the joint law of $k$ parallel draws
\citep{vilnis2023arithmetic,kool2019stochastic,parashar2025quasirandom},
bounded by \citet{li2026quasimotto} via a union-bound argument, and the two
axes compose.

\paragraph{Context-varying rules.} Context-varying rules
\citep{ants2026,su2026adaptivedecoding} select a prefix rule per context
from a policy rather than a fixed hyperparameter; treating that policy's
range as the family $\mathcal{G}$ in Corollary \ref{cor:grid} gives a
single scalar bounding every comparison the policy can induce, from one
decoding run under its pointwise-narrowest member.

\section{Prefix-Class Audit}
\label{app:audit}

The class assumption behind Table \ref{tab:rules} is confirmed on
$997{,}465{,}872$ real logged cutoffs, pooled across every decoding log
in our implementation's output (grid walks, pairwise walks and the $\rho$
sweep, not just the swept-family grid walks alone): zero
out-of-range cutoffs, zero non-nested pairs, among the rules implemented
here. Mirostat is not yet implemented, so its prefix-ness is inferred from
its published description rather than checked directly.

Top-W and locally typical sampling are the two rules outside our
guarantees. Top-W's cutoff follows a score $\varphi_i=f_i+\lambda\log p_i$
mixing token-embedding distance $f_i$ with log-probability
\citep{topw2026}, a different total order from the probability-sorted one
Definition \ref{def:prefix} requires, so its retained set is a prefix of
that order but not of ours. Its optimality argument yields a crop that is
a single token or a prefix of an order mixing probability with
token-embedding geometry, so Top-W, like locally typical sampling, lies
outside the prefix class.

Locally typical sampling can omit the top token, since
\citet{meister2023typical}'s rule sorts tokens by $|{-}\log p_{(i)}-H|$
ascending and keeps the shortest prefix of that order reaching cumulative
mass $\tau$, which need not include the token nearest probability $1$. At
$p=(0.4,0.3,0.3)$, $H\approx1.09$ nats, the two $0.3$ tokens sit closer to
$H$ than the $0.4$ token does, so at $\tau\le0.6$ the retained set is the
two $0.3$ tokens and the top token is dropped. Comparing $p_{(1)}$ against
$e^{-H}$ directly is not a measurement of this: since $-\log p_{(1)}$ is
the minimum surprisal over the support, $H=\Ex[-\log p_{(i)}]\ge-\log
p_{(1)}$ on every distribution, with equality only when every token with
positive mass is equally likely, so that comparison holds by construction
and says nothing about whether the mass-based rule retains the top token.
Implementing that rule and counting how often it omits the top token,
pooled over the seed-$0$ grid-walk log of each of the nine model/task
configurations of Section \ref{sec:experiments} ($2{,}798{,}071$ steps
total, per-configuration rates from $2.5\%$ to $24.3\%$), gives a pooled
omission rate of $8.8\%$ at $\tau=0.2$, a commonly used operating point for
this rule.

\section{Experimental Details}
\label{app:expdetails}

\paragraph{Calibration corpus.} The operating point $\rho(K)=\Ex[Z_K]$ is
measured over a teacher-forced calibration corpus of 200 held-out GSM8K
prompts, each extended with its known correct solution (the gold answer) as
the continuation, so the model's next-token distribution is logged against
real, on-distribution text rather than its own generations.

\paragraph{Computing a certificate.} Computing $\Sep_T(A,B)$ from a
decoding run follows directly from the coupling of Lemma \ref{lem:nest}:
for each prompt, walk $T$ steps under $A\wedge B$, at each step accumulate
$\log(Z_{\min}/Z_{\max})$ in log space over the retained masses of $A$ and
$B$, and average $1-e^{L}$ over prompts with a bootstrap interval. No
component beyond the retained masses and the pointwise-narrower draw is
needed, which is why the certificate is computable from any decoding run's
own logs.

\paragraph{Grid.} Each of the six tunable families is swept at six
log-spaced points, alternating $1\times$ and $3\times$ each power of ten:
top-$k\in\{10,20,40,80,160,320\}$, top-$p\in\{0.70,0.80,0.90,0.95,0.98,
0.99\}$, $\epsilon\in\{3{\times}10^{-4},1{\times}10^{-3},3{\times}10^{-3},
1{\times}10^{-2},3{\times}10^{-2},1{\times}10^{-1}\}$, $\eta$ over the same
six values, min-$p\in\{0.01,0.02,0.05,0.10,0.20,0.30\}$, and
top-$n\sigma\in\{0.5,0.8,1.0,1.5,2.0,3.0\}$. Basic (untruncated) and
$p$-less have no hyperparameter, and top-$H$ is used at its default
(top-$n{=}100$, $\alpha{=}0.4$) rather than swept, so the three contribute
one member each, giving $\mathcal{G}$ 39 members and $\binom{39}{2}=741$
pairs.

\paragraph{Batch size.} Implementation is a custom decoding loop over
HuggingFace Transformers, run at batch size one throughout, since a
batch-invariance check found next-token logits drifting by up to $0.94$
between batch size $1$ and batch sizes $8$ and $64$, so certificates are
batch-size dependent. Every estimator arm (coupled, prompt-paired,
independent) runs the same custom decoding loop at batch size one, and none
uses a batched path or vLLM \citep{kwon2023vllm}, so the batch-invariance
drift cannot have confounded any comparison in Section \ref{sec:results}.

\paragraph{Top-2048 exclusion.} Certificates use the top $2048$ logged
probabilities, and steps whose cutoff would exceed that are excluded
consistently across every certificate drawn from the same log. Excluding a
step drops a factor of at most $1$ from the retained-mass product rather
than the true, smaller factor, so this is anti-conservative in principle:
the size of the understatement depends on how far each excluded step's
true $r_t$ sits below $1$, not merely on how many steps are excluded, so
the excluded-step fraction alone is not itself a bound on the error. The
grid certificate needs no such exclusion at all. $\mathcal{G}^{\vee}$'s
cutoff is basic's, the untruncated rule, whose retained mass is exactly $1$
by construction regardless of how much of the tail is logged, so its true
value is used directly; summing only the logged head instead would
understate this mass by up to $9\%$ on the steps we checked, making any
certificate computed that way nominally anti-conservative.
$\mathcal{G}^{\wedge}$'s cutoff is $1$ on essentially every step, since
top-$k\!=\!10$ alone already bounds it at $10$, and $1$ or $10$ never
approaches the logged $2048$. An exclusion flag set whenever \emph{any} of
the grid's $39$ members needs the tail, rather than whenever
$\mathcal{G}^{\wedge}$ or $\mathcal{G}^{\vee}$ themselves do, would report
excluded-step fractions as high as $6.8\%$ on gpt-oss-20b/MATH-500; with
the check the grid certificate actually needs, no step is excluded on any
of the nine configurations.

\begin{table}[h]
\centering
\begin{threeparttable}
\caption{\textbf{At a matched operating point the certificate stays large and
stops being exact} (Qwen3-8B, MATH-500, $T{=}1024$). Rows are ordered by how
closely the pair was matched; the flip-rate paragraph above discusses the
dose-response reading of the flip-rate column.}
\label{tab:matched}
\small
\setlength{\tabcolsep}{7pt}
\renewcommand{\arraystretch}{1.10}
\begin{tabular}{@{}l S[table-format=1.3] S[table-format=1.3] S[table-format=2.2] S[table-format=1.3]@{}}
\toprule
& {achieved} & {} & {} & {} \\
\textbf{Pair, matched to $\rho=0.95$} & {$\rho$} & {$\Sep_{1024}$} &
{$\Delta_{1024}$} & {$\Pr[\text{flip}]$} \\
\midrule
min-$p$ vs top-$p$        & 0.954 & 0.993 & 13.67 & 0.998 \\
top-$n\sigma$ vs top-$p$  & 0.957 & 0.994 & 13.93 & 0.996 \\
$\eta$ vs top-$p$         & 0.978 & 0.998 & 27.55 & 0.374 \\
\bottomrule
\end{tabular}
\begin{tablenotes}[flushleft]\footnotesize
\item $\Pr[\text{flip}]$ is the fraction of trajectories on which the nesting
order flips; wherever it is positive the certificate is a bound, not an
equality. Single runs without intervals, and flip rate grows with $T$, so
this column is comparable only at $T{=}1024$. $\Sep$ at three decimals
fixes $1-\Sep$ to one significant figure, so compare rows by $\Delta$.
Qwen3-8B/MATH-500 is not one of the nine grid configurations, so it has no
grid certificate; the grid certificate never flips by construction
($\mathcal{G}^{\wedge}$ is narrower than $\mathcal{G}^{\vee}$ at every
context) and is therefore exact.
\end{tablenotes}
\end{threeparttable}
\end{table}

\paragraph{Compute.} All decoding runs to date total $64.5$
H200-GPU-hours, plus under one CPU-hour for prompt preparation and the
logs-only experiments, against the six thousand A100-hours of
\citet{schaeffer2025minp}, so certifying every family we tested cost under
two percent of one baseline's training-adjacent search.

\section{Additional Results}
\label{app:results}

\paragraph{Gate controls.} On Qwen3-1.7B/GSM8K ($T{=}256$, $n{=}200$, pair
min-$p{=}0.05$ versus top-$p{=}0.95$) every per-prompt first-step cutoff
matched exactly between an independent coupled read and an independent
solo read of the same prompt ($200/200$), the identical-pair control showed
zero divergence ($0/200$), and the greedy-versus-untruncated control
diverged on every prompt ($200/200$). The comparison against the point
prediction, that the certificate equals the probability a real simulated
coupled decoding run diverges, uses a percentile bootstrap over the
per-prompt divergence indicator with resampling unit prompt and $n{=}200$.

\begin{table}[h]
\centering
\caption{\textbf{Prompts needed to resolve a one-point difference with $k$
coupled samples per prompt} (default pair, GSM8K variance decomposition).
Benchmark sizes: GSM8K $1{,}319$, MATH-500 $500$,
GPQA-Diamond $198$, AIME 2025 $30$.}
\label{tab:benchmarks}
\small
\setlength{\tabcolsep}{8pt}
\begin{tabular}{@{}lll@{}}
\toprule
$k$ & $\tau^2=0$ (point estimate) & $\tau^2=9.3\times10^{-4}$ (upper bound) \\
\midrule
$1$         & $6{,}318$ & $6{,}392$ \\
$8$         & $790$     & $863$ \\
$32$        & $197$     & $271$ \\
$\approx210$ & within AIME 2025's $30$ (untested) & $103$ \\
any $k$     & no floor  & never below about $73$ \\
\bottomrule
\end{tabular}
\end{table}

\paragraph{The pairwise distribution over the grid.} The full pairwise
distribution over the grid's $741$ pairs is not yet reported. A pairwise
certificate can be read off a run under either member alone (Proposition
\ref{prop:onerule}), so the $741$ pairs need at most $39$ new single-rule
runs rather than $741$ pair-specific ones, cheaper but still undecoded. The
existing $\mathcal{G}^{\wedge}$-walked grid logs do not substitute for
this, since Proposition \ref{prop:onerule} needs a run under one specific
member, not under the grid's pointwise-narrowest rule, for which a
$\mathcal{G}^{\wedge}$ run remains not conservative. Of the $741$ grid
pairs, only the default one (min-$p$ versus top-$p$ at default
hyperparameters) has that log so far, separately from the
matched-operating-point pairs of Table \ref{tab:matched}: certificate
$0.945$, $95\%$ CI $[0.931,0.959]$ over $200$ prompts, computed by the
machinery that would produce the full distribution once the remaining $740$
pairs are decoded.

\paragraph{Why the flip-rate contrast is not established.} Three reasons
qualify the raw-versus-matched flip-rate contrast of Section
\ref{sec:results}. The raw and matched rates were measured on different
pairs and possibly different horizons, so the comparison is not paired.
Per-trajectory flip rate is monotone in $T$, so a horizon difference alone
could produce it. And within Table \ref{tab:matched} the flip rate is
monotone in matching quality, with the worst-matched pair flipping least,
consistent with the mechanism but also meaning the three points sit on a
dose-response curve at one operating point and all share top-$p$ as an arm.

\paragraph{Matched operating-point sweep.} On Qwen3-1.7B/GSM8K, min-$p$
versus top-$p$ at five matched operating points
$\rho\in\{0.82,0.87,0.90,0.95,0.98\}$, where $\rho{<}0.82$ is unreachable
for min-$p$ and $\eta$ on this model's real distributions since their
measured floors are $0.72$ and $0.80$, the per-step minority-sign rate is
$0.018$--$0.021$ and the trajectory flip rate $0.70$--$0.93$, against
$0.014$ and $0.99$ raw. The same sweep for $\eta$ versus min-$p$, the
matched pair not involving top-$p$, gives per-step rates $0.012$--$0.029$
and trajectory rates $0.67$--$0.98$. The correlation between
$\log(|{\Ex[g_t]}|/\mathrm{sd}[g_t])$ and the per-step minority-sign rate
across all eleven conditions, raw plus the ten matched, is $-0.53$.
Matching lowers the mean gap relative to its spread, in the direction
Equation \eqref{eq:lnd} predicts, but a correlation of this size across
eleven points is a moderate relationship rather than a clean collapse onto
one curve.

\paragraph{Long-horizon details.} The four AIME horizons are read from one
$T{=}32{,}768$ decoding run as true prefixes, so all four numbers come from
the same trajectories, and $\Sep_T$ is computed directly as
$1-\frac1n\sum_i e^{-S_i}$ over the $30$ prompts rather than from the
Jensen bound $1-e^{-\Delta_T}$. It is monotone non-decreasing in $T$, as
the theory requires once the same trajectories are reused rather than
resampled at each horizon. Generations do not all reach the cap:
$\mathrm{eos\_rate}$ (the fraction of the $30$ prompts finished by that
horizon) rises from $0$ at $T{=}1024$ to $0.07$, $0.47$ and $1.00$ at the
three longer caps, so we report $\Delta_T$ per realized token ($0.017$,
$0.015$, $0.016$, $0.016$ respectively) rather than per capped token, which
would understate the rate once most generations finish early. A
shorter-horizon run (Qwen3-1.7B/GSM8K, $T{=}256$ through $1024$) confirms
the methodology where every generation reaches the cap,
$\mathrm{eos\_rate}=0$. Table \ref{tab:matched} reports the
operating-point-matched certificates in full.

\paragraph{Coupling details.} The maximal coupling cuts variance
$3.8\times$ against prompt-level pairing on the default pair. The
shared-seed baseline and a shared-uniform-coupling ablation
\citep{vilnis2023arithmetic,li2026quasimotto} have $3.9\times$ and
$2.0\times$ the maximal coupling's variance on the same pair, both
intervals excluding $1$, so these two particular ways of
sharing randomness, not sharing randomness as such, fall short of its
reduction. Shared Gumbel-max noise, drawing one Gumbel vector per step and
taking each rule's argmax over its own retained set, is itself a
realization of the same maximal coupling for nested prefixes (the two
argmaxes agree with probability exactly $Z_A/Z_B$, verified against the
closed form to $4$ significant figures on a synthetic distribution), but
was not run here. The negative control has small coupled-arm outcome
variance ($0.040$) and a near-zero mean paired gap ($-0.002$), because both
rules answer most prompts the same way regardless of how often their
tokens diverge, not because the rules are similar as samplers; its
measured reduction is $(\sigma_A^2+\sigma_B^2)/\text{coupled variance}
\approx 12.0\times$, with $\sigma_A^2,\sigma_B^2$ measured on the
independent arms. For GPQA-Diamond the interval $[8.0,62.1]$ excludes $1$
while the bootstrap $p=0.16$ is not significant even before any correction
across the sweep. These need not agree: the interval is a percentile
interval of the bootstrap ratio itself, while the $p$-value tests against a
bootstrap distribution re-centered at its own mean, a different, more
conservative construction under skew, and this ratio's bootstrap
distribution is heavily right-skewed here. The reduction is therefore not
an artifact of the one saturated benchmark it was first measured on, since
MATH-500 replicates it with bootstrap $p=0.023$.

\paragraph{Predictiveness.} In Table \ref{tab:predictive}, a within-family
certificate as low as $0.17$ is not surprising: a monotone top-$k$ sweep
can reach $\Sep_T{=}0.10$, since nothing in the theory requires two members
of one family to be close. Under the null the measured gap's own standard
error scales with $\sqrt{\Sep_T}$, which by itself induces a positive
$\Sep_T$-versus-$|\text{gap}|$ correlation with no real prediction behind
it; standardizing each gap by its standard error is what that null noise
mechanism should remove but a real predictive relationship should not.
\texttt{code/predictiveness\_null\_check.py} reproduces both null
statistics.

\begin{figure}[h]
\centering
\begin{tikzpicture}[fig]
\begin{axis}[seplot, name=axG,
  width=9.4cm, height=6.6cm,
  xmode=log, xmin=3e-10, xmax=2e-2,
  xtick={1e-9,1e-8,1e-7,1e-6,1e-5,1e-4,1e-3,1e-2},
  xticklabels={$10^{-9}$,$10^{-8}$,$10^{-7}$,$10^{-6}$,$10^{-5}$,$10^{-4}$,$10^{-3}$,$10^{-2}$},
  xlabel={$1-\Sep_T(\mathcal{G})$, smaller means a tighter grid certificate},
  ymin=0.4, ymax=9.6, y dir=reverse,
  ytick={1,...,9},
  yticklabels={Qwen3-0.6B / GSM8K, Qwen3-1.7B / GSM8K, Qwen3-4B / GSM8K,
    Qwen3-8B / GSM8K, Qwen3-1.7B / GPQA-Diamond, Qwen3-1.7B / MATH-500,
    Qwen3-1.7B / AIME 2025, Gemma-3-4B / MATH-500, gpt-oss-20b / MATH-500},
  yticklabel style={font=\scriptsize, color=cInk},
  ymajorgrids=true,
  legend cell align=left, legend columns=3,
  legend style={at={(0.5,-0.22)}, anchor=north, row sep=-2pt,
                /tikz/every even column/.append style={column sep=8pt}},
  axis on top=false, clip=false]
  \addplot[cMute, line width=0.7pt, dash pattern=on 2pt off 1.5pt]
    coordinates {(1e-9,0.4) (1e-9,9.6)};
  \addlegendentry{floating-point floor, $10^{-9}$}
  \addplot[cRes!80, line width=0.7pt, dash pattern=on 2pt off 1.5pt]
    coordinates {(3e-4,0.4) (3e-4,9.6)};
  \addlegendentry{$\Sep_T = 0.9997$}
  \addplot[only marks, mark=*, mark size=2.2pt, cNarrow,
           mark options={fill=cNarrow, draw=white, line width=0.4pt},
           error bars/.cd, x dir=both, x explicit,
           error bar style={cNarrow, line width=0.6pt},
           error mark options={rotate=90, mark size=1.6pt, cNarrow, line width=0.6pt}]
    coordinates {
      (7.887e-6,1)  -= (1.640e-6,0)  += (2.032e-6,0)   
      (1.936e-3,2)  -= (2.320e-4,0)  += (2.320e-4,0)   
      (4.745e-5,3)  -= (1.480e-6,0)  += (1.470e-6,0)   
      (1.345e-5,4)  -= (3.726e-6,0)  += (3.720e-6,0)   
      (2.959e-4,5)  -= (1.970e-5,0)  += (1.380e-5,0)   
      (1.579e-9,6)  -= (5.130e-10,0) += (6.610e-10,0)  
      (1e-9,7)      -= (0,0)         += (0,0)          
    };
  \addlegendentry{five seeds, $95\%$ CI}
  \addplot[only marks, mark=diamond*, mark size=3pt, cWide,
           mark options={fill=cWide, draw=white, line width=0.4pt}]
    coordinates {(1.273e-4,8) (1e-9,9)};   
  \addlegendentry{single seed}
  \node[note, anchor=west, font=\tiny] at (axis cs:1.5e-9,7) {$<10^{-9}$};
  \node[note, anchor=west, font=\tiny] at (axis cs:1.5e-9,9) {$<10^{-9}$};
\end{axis}
\node[panel, anchor=south west] at ([xshift=-2.9cm,yshift=6pt]axG.north west)
  {Every grid certificate is at least $0.998$; eight of nine exceed $0.9997$};
\end{tikzpicture}
\caption{\footnotesize\textbf{The grid certificate on all nine
configurations,} as $1-\Sep_T(\mathcal{G})$ on a log axis, so a point
further right is a looser certificate. Circles are five-seed estimates with
$95\%$ CIs, diamonds single-seed estimates. Two values fall below
floating-point resolution and are drawn at $10^{-9}$. Only
Qwen3-1.7B/GSM8K ($0.998064$) lies right of the $\Sep_T=0.9997$ line.}
\label{fig:gridcert}
\end{figure}
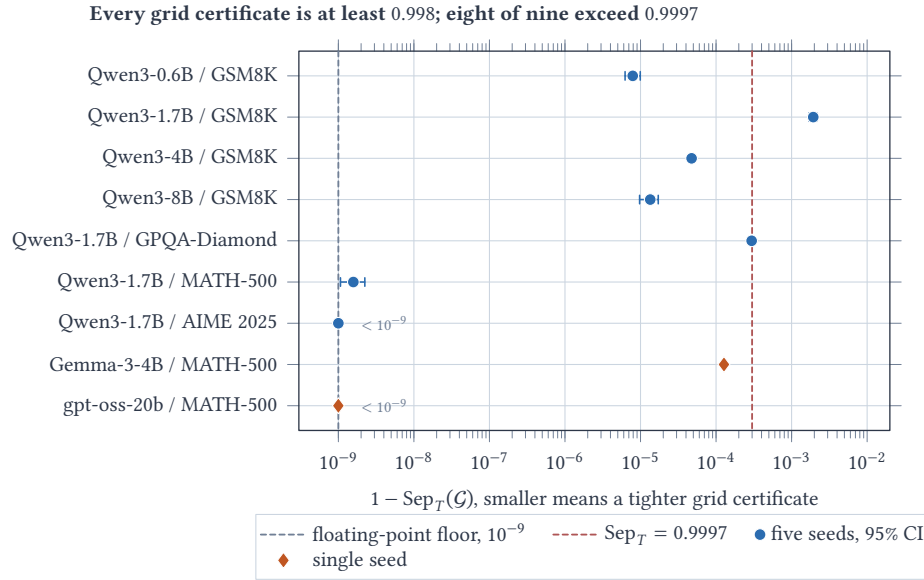

\begin{figure}[h]
\centering
\begin{tikzpicture}[fig]
\begin{axis}[seplot, name=axS,
  width=11.2cm, height=6.6cm,
  xmode=log, ymode=log,
  xmin=1, xmax=500, ymin=15, ymax=12000,
  xtick={1,2,5,10,20,50,100,200,500},
  xticklabels={$1$,$2$,$5$,$10$,$20$,$50$,$100$,$200$,$500$},
  ytick={30,100,300,1000,3000,10000},
  yticklabels={$30$,$100$,$300$,$1000$,$3000$,$10^4$},
  xlabel={coupled samples per prompt, $k$},
  ylabel={prompts needed to resolve a one-point gap},
  legend cell align=left, legend columns=3,
  legend style={at={(0.5,-0.22)}, anchor=north, row sep=-2pt,
                /tikz/every even column/.append style={column sep=6pt}},
  domain=1:500, samples=120]
  \addplot[draw=none, fill=cNarrow!7, forget plot]
    coordinates {(1,15) (5,15) (5,12000) (1,12000)} \closedcycle;
  \node[note, anchor=north east, align=right] at (axis cs:4.7,11000)
    {fitted on\\$k\le5$};
  \node[note, anchor=north west] at (axis cs:5.4,11000) {extrapolated};
  \addplot[name path=lo, draw=none, forget plot] {6318/x};
  \addplot[name path=hi, draw=none, forget plot] {6318/x+73.5};
  \addplot[fill=cWide!18, draw=none] fill between[of=lo and hi];
  \addlegendentry{to bootstrap upper bound of $\tau^2$}
  \addplot[cInk, line width=0.9pt] {6318/x};
  \addlegendentry{point estimate, $\tau^2=0$}
  \addplot[cWide, line width=0.6pt, dash pattern=on 1.5pt off 1.5pt] {6318/x+73.5};
  \addlegendentry{upper bound: floor $\approx73$, never reaches $30$}
  \addplot[cMute, line width=0.5pt, dash pattern=on 2.5pt off 1.5pt, forget plot]
    coordinates {(1,1319) (500,1319)};
  \node[lbl, anchor=south east, color=cMute] at (axis cs:470,1350) {GSM8K, 1{,}319};
  \addplot[cMute, line width=0.5pt, dash pattern=on 2.5pt off 1.5pt, forget plot]
    coordinates {(1,500) (500,500)};
  \node[lbl, anchor=south east, color=cMute] at (axis cs:470,512) {MATH-500, 500};
  \addplot[cMute, line width=0.5pt, dash pattern=on 2.5pt off 1.5pt, forget plot]
    coordinates {(1,198) (500,198)};
  \node[lbl, anchor=south east, color=cMute] at (axis cs:470,203) {GPQA-Diamond, 198};
  \addplot[cMute, line width=0.5pt, dash pattern=on 2.5pt off 1.5pt, forget plot]
    coordinates {(1,30) (500,30)};
  \node[lbl, anchor=south west, color=cMute] at (axis cs:1.06,31) {AIME 2025, 30};
  \addplot[only marks, mark=*, mark size=2.0pt, mark options={fill=cInk, draw=white}]
    coordinates {(4.79,1319) (12.6,500) (31.9,198) (211,30)};
  \addlegendentry{point estimate crosses at $k$}
  \node[lbl, anchor=north east] at (axis cs:4.6,1240) {$k{=}4.8$};
  \node[lbl, anchor=north east] at (axis cs:12.2,470) {$k{=}12.6$};
  \node[lbl, anchor=north east] at (axis cs:30.5,186) {$k{=}31.9$};
  \node[lbl, anchor=north east] at (axis cs:200,28.5) {$k{=}211$};
  \addplot[only marks, mark=o, mark size=2.0pt, cWide, line width=0.6pt]
    coordinates {(5.07,1319) (14.8,500) (50.7,198)};
  \addlegendentry{upper bound crosses at $k$}
  \node[lbl, anchor=south west, color=cWide] at (axis cs:5.7,1420) {$5.1$};
  \node[lbl, anchor=south west, color=cWide] at (axis cs:16.5,540) {$14.8$};
  \node[lbl, anchor=north west, color=cWide] at (axis cs:56,188) {$50.7$};
\end{axis}
\node[panel, anchor=south west] at ([yshift=3pt]axS.north west)
  {Resampling per prompt removes the constraint for the larger benchmarks};
\end{tikzpicture}
\caption{\textbf{Prompts needed to resolve a
one-point difference with $k$ coupled samples per prompt} (default pair,
GSM8K variance; Table \ref{tab:benchmarks}). The solid line is the point
estimate $\tau^2=0$; the band reaches the upper bound of $\tau^2$, where the
requirement never falls below about $73$. Markers show where each curve
meets a benchmark's size (dashed). The model is fitted on five samples per
prompt, so everything beyond $k=5$ is extrapolated, with GSM8K's $\tau^2$
used for every benchmark.}
\label{fig:samplesize}
\end{figure}
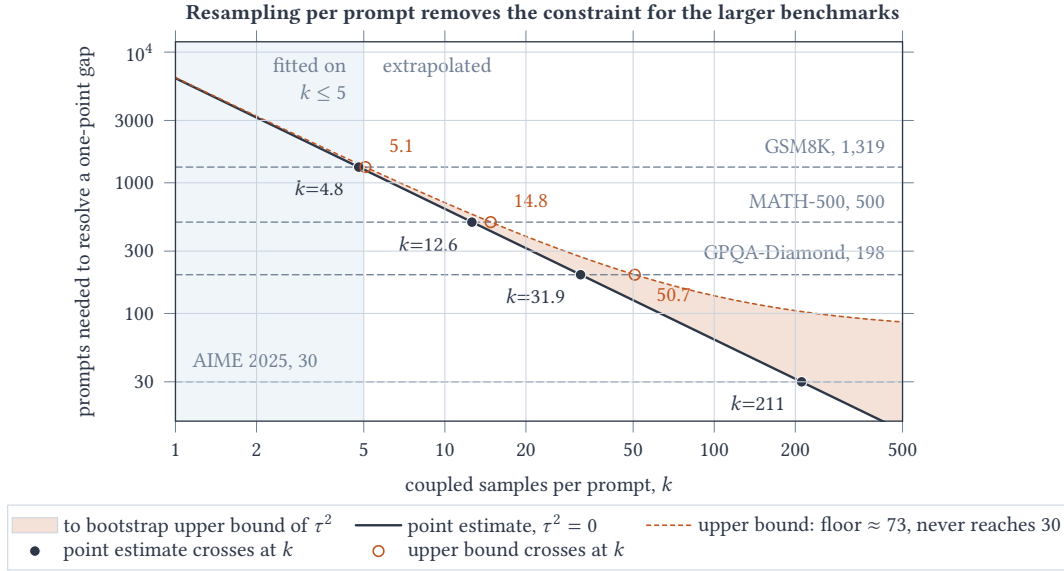

\paragraph{Benchmark-size derivation.} The $k$-sample requirements of
Table \ref{tab:benchmarks} decompose the five-seed coupled outcomes for the
default pair on GSM8K into within- and between-prompt variance
($\sigma_w^2$ and $\tau^2$); Appendix \ref{app:verification} describes the
script that reproduces them from the raw per-seed outcomes. The $k{=}8$
point-estimate figure, $790$, is close to but not the same number as
GPQA-Diamond's own $k{=}1$ coupled requirement ($787$): the two are
unrelated quantities, computed from different pairs' variance
decompositions, that happen to be numerically close. At the upper
confidence bound, the required count never falls below about $73$ however
large $k$ is taken, since $\tau^2$ alone then accounts for more than AIME
2025's $30$ prompts can resolve.

\begin{table}[h]
\centering
\begin{threeparttable}
\caption{\textbf{The eight pairs behind the predictiveness check}
(Qwen3-1.7B, GSM8K, $T{=}256$, five seeds). Family marks whether the pair
compares two different truncation rule types (cross) or the same rule at
two hyperparameter settings (within), including the negative control.}
\label{tab:predictive}
\small
\setlength{\tabcolsep}{6pt}
\renewcommand{\arraystretch}{1.10}
\begin{tabular}{@{}l l S[table-format=1.3] S[table-format=-1.3] l@{}}
\toprule
\textbf{Pair} & \textbf{Family} & {$\Sep_T$} & {gap} & {gap 95\% CI} \\
\midrule
min-$p$(0.05) vs top-$p$(0.95)   & cross  & 0.710 & 0.017  & $[0.000,0.034]$ \\
top-$k$(10) vs top-$p$(0.70)     & cross  & 0.992 & -0.023 & $[-0.052,0.006]$ \\
$\epsilon$(0.1) vs $\eta$(0.0003)& cross  & 0.989 & 0.013  & $[-0.015,0.041]$ \\
top-$p$(0.98) vs top-$p$(0.99)   & within & 0.437 & -0.002 & $[-0.014,0.010]$ \\
top-$k$(10) vs top-$k$(320)      & within & 0.541 & 0.003  & $[-0.009,0.015]$ \\
top-$k$(10) vs top-$k$(80)       & within & 0.517 & 0.003  & $[-0.008,0.014]$ \\
top-$k$(80) vs top-$k$(320)      & within & 0.174 & 0.000  & $[-0.004,0.004]$ \\
min-$p$(0.01) vs min-$p$(0.30)   & within & 0.992 & 0.013  & $[-0.015,0.041]$ \\
\bottomrule
\end{tabular}
\begin{tablenotes}[flushleft]\footnotesize
\item gap is the mean signed accuracy difference over $1{,}000$
coupled-arm observations ($200$ prompts $\times$ five seeds); the
certificate bounds $|\text{gap}|$. Observations lie in $\{-1,0,1\}$, so
means and bootstrap percentiles are multiples of $0.001$, which is why two
rows share gap and CI ($0.013$, $[-0.015,0.041]$) although their $\Sep_T$
and mean absolute gaps ($0.203$ vs $0.205$) differ. The default pair's
$0.710$ differs from the gate's $0.722$ at $T{=}256$ by Monte Carlo
variation across decoding sessions, within the gate's interval. These
certificates are computed online against the full softmax, so no top-$2048$
exclusion applies.
\end{tablenotes}
\end{threeparttable}
\end{table}

\section{Mechanical Verification of the Theory}
\label{app:verification}

The released code checks every formal claim numerically on synthetic
models, and is written to falsify rather than confirm. It verifies prefix
membership and nesting for the swept rules of Table \ref{tab:rules}
(including ties and boundary cases), Lemma \ref{lem:nest}(i) to machine
precision and the marginals and maximality of Lemma \ref{lem:nest}(ii) by
Monte Carlo, and the bounds of Proposition \ref{prop:var} against
adversarially constructed outcomes. Exact enumeration over all sequences
of context-dependent synthetic models confirms that the certificate's
expectation must be taken along $A\wedge B$ (substituting
$\mathcal{G}^{\wedge}$ is not conservative), that $\Sep_T$ equals the
sequence-level total variation exactly on non-flipping pairs and exceeds
it otherwise, the equality condition of Equation \eqref{eq:lnd}, the
dominance of the grid certificate over every pairwise certificate,
Proposition \ref{prop:onerule}, the attaining metric of Appendix
\ref{app:extras}, the Gumbel-max agreement rate $Z_A/Z_B$, and the locally
typical counterexample of Appendix \ref{app:audit}. A second script checks
Proposition \ref{prop:fragile} against its exact three-state computation
and both parts of Corollary \ref{cor:budget} against the exact
distribution of $S$. All $28$ checks pass, and a deliberately incorrect
variant of Equation \eqref{eq:lnd} fails as it should. These checks show
that theory and code agree on synthetic inputs; no number from them is
reported as a property of a language model.

Two real-model computations have their own scripts.
\texttt{code/benchmark\_size\_analysis.py} reproduces the benchmark-size
requirements and the within/between-prompt decomposition with its
bootstrap upper bound on $\tau^2$, and
\texttt{code/aime\_horizons\_as\_prefixes.py} computes the four AIME
certificates as prefixes of one $T{=}32{,}768$ run, along which $\Sep_T$ is
monotone in $T$.

\end{document}

%% file: affiliations.tex
\DeclareAffiliation{hbku}{%
  College of Science and Engineering, Hamad Bin Khalifa University, Doha, Qatar}

\DeclareAffiliation{tamu}{%
  Department of Computer and Electrical Engineering,
  Texas A\&M University, College Station, TX, USA}

\DeclareAffiliation{iub}{%
  Luddy School of Informatics, Computing, and Engineering,
  Indiana University Bloomington, Bloomington, IN, USA}


%% file: figures/sep_fig_teaser.tex
\begin{tikzpicture}[fig, x=1cm, y=1cm]

\begin{scope}
\def\bs{0.36}\def\bw{0.26}\def\hs{11.0}\def\ph{3.3}
\def\KA{4}\def\KB{9}\def\NB{16}
\fill[cNarrow!9] (0,0) rectangle (\KA*\bs,\ph);
\fill[cBand!24]  (\KA*\bs,0) rectangle (\KB*\bs,\ph);
\foreach \i/\p in {1/0.28,2/0.19,3/0.13,4/0.095}{
  \draw[barboth] ({(\i-1)*\bs+0.055},0) rectangle ({(\i-1)*\bs+0.055+\bw},{\p*\hs});}
\foreach \i/\p in {5/0.072,6/0.055,7/0.042,8/0.032,9/0.025}{
  \draw[barhi]   ({(\i-1)*\bs+0.055},0) rectangle ({(\i-1)*\bs+0.055+\bw},{\p*\hs});}
\foreach \i/\p in {10/0.019,11/0.015,12/0.012,13/0.010,14/0.008,15/0.0065,16/0.005}{
  \draw[bardrop] ({(\i-1)*\bs+0.055},0) rectangle ({(\i-1)*\bs+0.055+\bw},{\p*\hs});}
\draw[rule] (0,0) -- (\NB*\bs+0.12,0);
\draw[rule] (0,0) -- (0,\ph+0.08);
\node[lbl, rotate=90, anchor=south] at (-0.12,\ph/2) {$p_{(i)}$, sorted};
\node[lbl, anchor=north] at ({0.055+\bw/2},-0.04) {$i{=}1$};
\node[lbl, anchor=north] at ({(\NB-1)*\bs+0.055+\bw/2},-0.04) {$V$};
\node[note, anchor=north] at ({(\NB-3.3)*\bs},-0.06) {$\cdots$};
\draw[cut, cNarrow] (\KA*\bs,0) -- (\KA*\bs,\ph+0.08)
  node[lbl, above, color=cNarrow] {$K_A$};
\draw[cut, cWide]   (\KB*\bs,0) -- (\KB*\bs,\ph+0.08)
  node[lbl, above, color=cWide] {$K_B$};
\node[lbl, align=center, color=cWide, font=\scriptsize\bfseries]
  at ({(\KA+\KB)/2*\bs},1.15) {disputed\\band};
\draw[decorate, decoration={brace, mirror, amplitude=3pt}, cNarrow, line width=0.5pt]
  (0.03,-0.42) -- (\KA*\bs-0.03,-0.42)
  node[midway, below=2.5pt, lbl, color=cNarrow] {$Z_A$};
\draw[decorate, decoration={brace, mirror, amplitude=3pt}, cWide, line width=0.5pt]
  (0.03,-0.80) -- (\KB*\bs-0.03,-0.80)
  node[midway, below=2.5pt, lbl, color=cWide] {$Z_B$};
\node[lbl, align=left, anchor=north east, draw=cGrid, fill=white,
      rounded corners=2pt, inner sep=3pt, line width=0.4pt]
  at (\NB*\bs+0.12,\ph+0.04) {%
  $P_A = P_B(\,\cdot \mid S_A)$\\
  agree w.p.\ $r = Z_A/Z_B$\\
  $\TV(P_A,P_B) = 1 - r$};
\draw[barboth] (0,-1.55) rectangle ++(0.22,0.17);
\node[lbl, anchor=west] at (0.26,-1.465) {$S_A$: kept by both};
\draw[barhi]   (2.72,-1.55) rectangle ++(0.22,0.17);
\node[lbl, anchor=west] at (2.98,-1.465) {band};
\draw[bardrop] (3.82,-1.55) rectangle ++(0.22,0.17);
\node[lbl, anchor=west] at (4.08,-1.465) {dropped by both};
\node[panel, anchor=south west] at (-0.42,\ph+0.55)
  {(a)\enspace One step: two prefix rules are nested};
\end{scope}

\begin{scope}[shift={(6.65,0)}]
\def\T{4.25}\def\ph{3.3}
\foreach \y/\xf in {3.3/1.45, 2.72/3.15, 1.56/0.75, 0.40/2.35}{
  \draw[draw=cInk!55, line width=1.1pt] (0,\y) -- (\xf,\y);
  \draw[draw=cNarrow, line width=0.9pt] (\xf,\y) -- ++(0.32,0.15) -- (\T,{\y+0.15});
  \draw[draw=cWide,   line width=0.9pt] (\xf,\y) -- ++(0.32,-0.15) -- (\T,{\y-0.15});
  \fill[cBand, draw=cWide, line width=0.5pt] (\xf,\y) circle (1.7pt);
  \node[lbl, anchor=west] at (\T+0.14,\y) {$D_i \in \{-1,0,1\}$};
}
\foreach \y in {2.14, 0.98}{
  \draw[draw=cInk!55, line width=1.1pt] (0,\y) -- (\T,\y);
  \fill[cInk!55] (\T,\y) circle (1.2pt);
  \node[lbl, anchor=west, color=cNarrow] at (\T+0.14,\y) {$D_i = 0$ exactly};
}
\draw[rule] (0,-0.05) -- (\T+0.10,-0.05);
\draw[cut] (\T,-0.05) -- (\T,\ph+0.14) node[lbl, above=1pt] {$T$};
\node[lbl, anchor=north] at (0,-0.08) {$t{=}1$};
\node[lbl, anchor=north] at (\T/2,-0.08) {decoding step $t$};
\node[lbl, rotate=90, anchor=south] at (-0.12,\ph/2+0.15) {prompts $i=1,\dots,6$};
\node[lbl, align=left, anchor=north west] at (-0.42,-0.55) {%
  $\Sep_T = \Pr[\text{fork by } T]$ bounds $|\Ex f(Y^A) - \Ex f(Y^B)|$;
  here $4$ of $6$ prompts fork.\\
  Unforked prompts give $D_i = 0$ exactly and add no variance.};
\draw[draw=cInk!55, line width=1.1pt] (-0.15,-1.465) -- ++(0.32,0);
\node[lbl, anchor=west] at (0.21,-1.465) {shared prefix ($A\wedge B$)};
\draw[draw=cNarrow, line width=0.9pt] (3.30,-1.465) -- ++(0.32,0);
\node[lbl, anchor=west] at (3.66,-1.465) {rule $A$};
\draw[draw=cWide, line width=0.9pt] (4.80,-1.465) -- ++(0.32,0);
\node[lbl, anchor=west] at (5.16,-1.465) {rule $B$};
\fill[cBand, draw=cWide, line width=0.5pt] (6.40,-1.465) circle (1.7pt);
\node[lbl, anchor=west] at (6.53,-1.465) {band draw};
\node[panel, anchor=south west] at (-0.42,\ph+0.55)
  {(b)\enspace Many steps: coupled pairs fork only on a band draw};
\end{scope}

\end{tikzpicture}

%% file: figures/sep_fig_theory.tex
\begin{tikzpicture}[fig]

\begin{axis}[seplot, name=axA,
  width=6.4cm, height=6.2cm,
  xmode=log, ymode=log,
  xmin=600, xmax=70000, ymin=0.03, ymax=300000,
  xtick={1024,4096,16384,32768},
  xticklabels={$2^{10}$,$2^{12}$,$2^{14}$,$2^{15}$},
  ytick={0.1,1,10,100,1000,10000,100000},
  yticklabels={$0.1$,$1$,$10$,$100$,$10^{3}$,$10^{4}$,$10^{5}$},
  xlabel={horizon $T$ (tokens)},
  ylabel={$\Delta_T=\Ex\sum_{t\le T}|g_t|$},
  legend pos=north west, legend cell align=left,
  legend style={row sep=-2pt}]
  \addplot[draw=none, fill=cRes!7, forget plot]
    coordinates {(600,0.10536) (70000,0.10536) (70000,300000) (600,300000)} \closedcycle;
  \addplot[cRes, line width=0.8pt] coordinates {(600,0.10536) (70000,0.10536)};
  \addlegendentry{budget for $\Sep_T\le0.1$, $-\log0.9$}
  \node[note, anchor=north west, color=cRes!80!black] at (axis cs:640,0.095)
    {shaded: budget not met};
  \addplot[cMute, line width=0.5pt, dash pattern=on 1.5pt off 1.5pt,
           domain=600:70000, samples=40] {0.016*x};
  \addlegendentry{$0.016\,T$: AIME per-token rate}
  \addplot[cWide, line width=0.8pt, mark=*, mark size=1.9pt,
           mark options={fill=cWide, draw=cWide}]
    coordinates {(1024,17.2) (4096,59.7) (16384,202.8) (32768,256.2)};
  \addlegendentry{min-$p$ vs top-$p$, AIME 2025}
  \addplot[only marks, mark=triangle*, mark size=2.4pt,
           mark options={fill=cNarrow, draw=cNarrow}]
    coordinates {(1024,13.67) (1024,13.93) (1024,27.55)};
  \addlegendentry{matched pairs, MATH-500}
  \node[lbl, anchor=south west, align=left, draw=cGrid, fill=white,
        inner sep=2.5pt, line width=0.4pt] at (axis cs:640,0.135) {%
    over the $\Sep_T\le0.1$ budget:\\
    \textcolor{cNarrow}{matched pairs $130$--$261\times$}\\
    \textcolor{cWide}{AIME $163\times$ to $2431\times$}};
\end{axis}
\node[panel, anchor=south west] at ([yshift=3pt]axA.north west)
  {(a)\enspace Budget is flat in $T$; measured $\Delta_T$ grows};

\begin{axis}[seplot, name=axB, at={(axA.east)}, anchor=west, xshift=1.6cm,
  width=6.4cm, height=6.2cm,
  ymode=log,
  xmin=0, xmax=1100, ymin=1e-6, ymax=3,
  xtick={0,100,250,500,750,1000},
  ytick={1e-6,1e-5,1e-4,1e-3,1e-2,1e-1,1},
  yticklabels={$10^{-6}$,$10^{-5}$,$10^{-4}$,$10^{-3}$,$10^{-2}$,$10^{-1}$,$1$},
  xlabel={horizon $T$ (tokens)},
  ylabel={$\Pr[\text{no flip through } T]$},
  legend pos=south west, legend cell align=left,
  legend style={row sep=-2pt},
  domain=1:1100, samples=240]
  \draw[cut, cInk!60] (axis cs:100,1e-6) -- (axis cs:100,3);
  \draw[cut, cInk!60] (axis cs:1000,1e-6) -- (axis cs:1000,3);
  \addplot[cNarrow, line width=0.9pt] {1};
  \addlegendentry{within family, $\varepsilon{=}0$: exact}
  \addplot[cInk, line width=0.7pt] {2*(1-0.001)^x-(1-0.002)^x};
  \addlegendentry{bound, $\varepsilon{=}0.001$}
  \addplot[cInk, line width=0.7pt, dash pattern=on 3pt off 1.5pt] {2*(1-0.003)^x-(1-0.006)^x};
  \addlegendentry{bound, $\varepsilon{=}0.003$}
  \addplot[cInk, line width=0.7pt, dash pattern=on 1.2pt off 1.2pt] {2*(1-0.01)^x-(1-0.02)^x};
  \addlegendentry{bound, $\varepsilon{=}0.01$}
  \addplot[only marks, mark=triangle*, mark size=2.4pt,
           mark options={fill=cNarrow, draw=cNarrow}]
    coordinates {(1024,0.626) (1024,0.004) (1024,0.002)};
  \addlegendentry{matched pairs, $1-\Pr[\text{flip}]$}
\end{axis}
\node[panel, anchor=south west] at ([yshift=3pt]axB.north west)
  {(b)\enspace Exactness decays once the order can flip};

\end{tikzpicture}